\documentclass{article} 
\usepackage{siunitx}
\usepackage{caption}
\usepackage{subcaption}
\usepackage{tikz}
\usepackage{hyperref}

\usepackage{helvet}

\usepackage[utf8]{inputenc}

\usepackage{dcolumn}           
\newcolumntype{.}   {D{.}{.}{-1}} 
\newcolumntype{d}[1]{D{.}{.}{#1}} 
\newcolumntype{e}   {D{E}{E}{-1}} 
\newcolumntype{E}[1]{D{E}{E}{#1}} 

\usepackage{amsmath}
\DeclareMathSizes{7}{7}{3}{3} 
\usepackage{pifont}
\usepackage{amsfonts}
\usepackage{amssymb}
\usepackage{amsthm}

\usepackage{float}

\usetikzlibrary{decorations.pathreplacing} 
\usetikzlibrary{arrows,decorations.markings} 
\usetikzlibrary{patterns}
\def\hexagonsize{0.3cm}
\pgfdeclarepatternformonly
  {hexagons}
  {\pgfpointorigin}
  {\pgfpoint{3*\hexagonsize}{0.866025*2*\hexagonsize}}
  {\pgfpoint{3*\hexagonsize}{0.866025*2*\hexagonsize}}
  {
   \pgfsetlinewidth{0.4pt}
   \pgftransformshift{\pgfpoint{0mm}{0.866025*\hexagonsize}}
   \pgfpathmoveto{\pgfpoint{0mm}{0mm}}
   \pgfpathlineto{\pgfpoint{0.5*\hexagonsize}{0mm}}
   \pgfpathlineto{\pgfpoint{\hexagonsize}{-0.866025*\hexagonsize}}
   \pgfpathlineto{\pgfpoint{2*\hexagonsize}{-0.866025*\hexagonsize}}
   \pgfpathlineto{\pgfpoint{2.5*\hexagonsize}{0mm}}
   \pgfpathlineto{\pgfpoint{3*\hexagonsize+0.2mm}{0mm}}
   \pgfpathmoveto{\pgfpoint{0.5*\hexagonsize}{0mm}}
   \pgfpathlineto{\pgfpoint{\hexagonsize}{0.866025*\hexagonsize}}
   \pgfpathlineto{\pgfpoint{2*\hexagonsize}{0.866025*\hexagonsize}}
   \pgfpathlineto{\pgfpoint{2.5*\hexagonsize}{0mm}}
   \pgfusepath{stroke}
  }

\title{Comparison of Innovative Strategies for the Coverage Problem: Path Planning, Search Optimization, and Applications in Underwater Robotics}

\author{Ahmed Ibrahim $^{1,2}$, Francisco F. C. Rego $^{3,4,5}$ and Éric Busvelle $^{2}$}

\begin{document}

\maketitle
\begin{abstract}
In many applications, including underwater robotics, the coverage problem requires an autonomous vehicle to systematically explore a defined area while minimizing redundancy and avoiding obstacles. This paper investigates coverage path planning strategies to enhance the efficiency of underwater gliders, particularly in maximizing the probability of detecting a radioactive source while ensuring safe navigation.\\
We evaluate three path-planning approaches: the Traveling Salesman Problem (TSP), Minimum Spanning Tree (MST), and Optimal Control Problem (OCP). Simulations were conducted in MATLAB, comparing processing time, uncovered areas, path length, and traversal time. Results indicate that OCP is preferable when traversal time is constrained, although it incurs significantly higher computational costs. Conversely, MST-based approaches provide faster but less optimal solutions. These findings offer insights into selecting appropriate algorithms based on mission priorities, balancing efficiency and computational feasibility.
\end{abstract}

\footnotetext{
$^{1}$ \quad Instituto Superior Técnico, University of Lisbon, Lisbon, Portugal;\\
$^{2}$ \quad Université de Toulon, Aix Marseille Univ, CNRS, LIS, Marseille, France; busvelle@univ-tln.fr;\\
$^{3}$ \quad COPELABS from the Lusófona University, 
Lisbon, Portugal;\\
$^{4}$ \quad CTS - Centro de Tecnologia e Sistemas, UNINOVA-Instituto de Desenvolvimento de Novas Tecnologias, Caparica, Portugal;\\
$^{5}$ \quad Intelligent Systems Associate LAboratory (LASI), Portugal
}

	\section{Introduction}
\label{chap:intro}

Locating specific targets in an environment often requires a trajectory generation algorithm to ensure a high probability of detection. This leads us to an important concept which is motion planning or path planning. Geographic information systems, robotics, computer graphics, and other industries all encounter path-planning or motion-planning problems. The problem consists of finding an optimal route that avoids obstacles and achieves a certain goal. Normally, the path's quality is evaluated based on its Euclidean path length, smoothness, and obstacle-free status. In particular, the coverage problem is considered, which amounts to visiting every point in a defined area within a predefined distance.

Throughout this paper, several covering algorithms are to be discussed and implemented based on the state-of-the-art planning methods and coverage problem mentioned in chapter \ref{chap:literature}. These methods can be classified into standard path planning methods e.g. Traveling Salesman Problem (TSP), Minum Sppanning Tree (MST), etc., and search theory using Optimal Control Problems (OCP) as in the Koopman Search Theory\cite{koopman_theory_1956}. Chapter \ref{chap:Model} discusses the theory behind TSP, MST, and OCP. Also, it shows a framework for modeling basic search problems that were established as part of WWII antisubmarine warfare activities and have since been widely used in industrial and tactical applications \cite{chudnovsky2023search}. Afterwards, the numerical methods that are now available to solve the problem at hand are discussed. 

Several implemented algorithms can be found in Section~\ref{chap:Results} associated with multiple simulations and graphs, and a comparison between the aforementioned algorithms is done according to several key points e.g. Length of the traversed path and time taken during each path. For, the MST problem, two configurations were utilised, which are a square configuration and a hexagonal configuration, to define the path traversed in each algorithm.


\section{Literature Review}\label{chap:literature}
\subsection{Search Theory Problem}
Search theory \cite{chung2011search} plays a crucial role in the field of robotics, offering a structured approach to solving problems related to navigation, decision-making, and resource management. One of its primary applications is path planning, particularly for autonomous robots. By employing search theory, robots can determine the most efficient route to maximize the probability of detecting a determined object while avoiding obstacles and ensuring safe navigation through complex environments.

Moreover, search theory enables the optimization of resources in robotics. Robots often have limited resources, such as energy and computational power. Through the principles of search theory, they can plan their actions to minimize energy consumption and maximize operational time. This is of paramount importance in scenarios where robots are deployed in remote or hostile environments and must operate efficiently with constrained resources.


Furthermore, search theory is useful in exploration and mapping activities. When robots are entrusted with mapping unknown or partially known settings, this theory guides their exploration, allowing them to efficiently map the surroundings and uncover new areas or places of interest. It is essential in applications such as search and rescue, environmental monitoring, and archaeological exploration.


Starting with \cite{chung2011search}, which covers recent advances in pursuit-evasion and autonomous search that apply to mobile robotics applications. Karaman \& Frazzoli (2011) \cite{10.1177/0278364911406761} discuss sampling-based algorithms for optimal motion planning in robotics. They present algorithms such as Probabilistic Roadmaps (PRM) and Rapidly-exploring Random Trees (RRT) that have been shown to work well in practice and have theoretical guarantees of probabilistic completeness. Masakuna and Fukuda Masakuna et al. (2019) \cite{10.1109/irc.2019.00086}  address the problem of search by a group of solitary robots. They propose a coordinated search strategy for self-interested robots without prior knowledge about each other. The results demonstrate the effectiveness of the strategy in achieving efficient search. Ismail \& Hamami (2021) \cite{10.3390/app11052383}  conduct a systematic literature review on swarm robotics strategies applied to the target search problem with environment constraints. The review identifies various strategies and highlights their applicability in different research domains, including computational and swarm intelligence.


\subsection{Coverage Problem}

For several convincing reasons, the coverage problem is extremely important in the realm of robotics. First and foremost, it is critical to efficiently use resources, particularly when resources such as time, energy, or materials are limited. To properly complete their tasks, robots must make well-informed decisions on how to navigate and cover an area. Several researchers consider the covering problem as part of the motion planning problem.

A critical component related to the coverage issue is time efficiency. Identifying the most time-efficient methods for area coverage is critical in search and rescue missions, inspection activities, and surveillance, especially when time is of the essence. Furthermore, in mobile robotic systems, energy is frequently a valuable and restricted resource. Efficient coverage tactics are critical for reducing energy consumption, prolonging the robot's operational time, and improving overall efficacy.

The coverage problem guarantees that sensors mounted on robots acquire information equally across the entire region, lowering the danger of missing key data points in data-collecting applications such as environmental monitoring and exploration. In other cases, such as floor cleaning robots or crop harvesting equipment, complete coverage is required because missing any part of the area is undesirable.

Furthermore, coverage methods are frequently included in path planning. They aid in the generation of pathways that not only cover the area effectively but also take obstacle avoidance and safe navigation into account, especially in dynamic contexts. The coverage problem has several practical applications, including autonomous lawnmowers, vacuum cleaners, agricultural robots, and autonomous cars, all of which require good area coverage to complete their tasks.

The coverage problem in robotics refers to the task of ensuring that one or more robots visit each point in a target area at least once. Several papers have addressed this problem and proposed different approaches and algorithms.

Batalin \& Sukhatme (2004) \cite{10.1023/b:tels.0000029038.31947.d1} defined the coverage problem as the maximization of the total area covered by a robot's sensors. They discussed the exploration and deployment of a mobile robot in a communication network and proposed algorithms for efficient coverage.

Galceran \& Carreras (2013) \cite{10.1016/j.robot.2013.09.004} conducted a survey on coverage path planning for robotics. They presented a collection of algorithms for complete coverage path planning using a team of mobile robots in unknown environments. They discussed various approaches, including boustrophedon decomposition and cellular decomposition.\\
Algorithms mentioned by Batalin \& Sukhatme (2004) \cite{10.1023/b:tels.0000029038.31947.d1} and Le et al. \cite{10.3390/s18082585} improved the efficiency of coverage path planning. However, resource constraints and coverage optimization were not studied, which was done by Strimel \& Veloso (2014) \cite{10.1109/iros.2014.6942969}.

\subsection{Minimum Spanning Tree}

A Minimum Spanning Tree (MST) is a graph tree given by a subset of edges that connects all nodes while minimizing the overall sum of edge weights. In robotics, these nodes can represent critical sites that the robot must visit, and the MST acts as a roadmap for linking these spots effectively. Robots can cross the MST's edges to visit all essential places while traveling the least amount of distance, see \cite{4392963}.

The spanning-tree coverage has already been described in
\cite{gabriely2001spanning}
and in
\cite{hazon2005redundancy}.
One of the main interests of this method is its cost in terms of computation time, which is negligible compared to approaches based on approximate resolutions of integer optimization problems, such as the traveling salesman problem (polynomial cost, compared to an exponential cost for the exact TSP algorithm).

This approach can be used for a single robot or for a fleet of N robots. In the latter case, there are several ways to assign the surface to be explored to each robot:
\begin{itemize}
\item The surface can be divided into N areas of equal surface a priori, but this does not guarantee optimality since some areas can be much more difficult to explore;
\item the problem can be posed and solved for a single robot, then the trajectory will be divided into N trajectories of equal length, which does not allow to recover the robots at the place from where they started;
\item Finally, the calculated minimum spanning tree can be divided into a forest of trees (by removing some branches) of the same size, which then allows to recover N closed trajectories. If N=2, both robots will be released and recovered at the same point.
\end{itemize}

There are many articles on the optimal exploration of a terrain with one or more robots. Some are more specifically devoted to the study of this problem for marine robots. In \cite{merci2025management}, the authors particularly study the transition to real conditions by focusing on practical constraints. A real experiment made it possible to validate the approach on four gliders.
In \cite{li2024simulated}, the spanning tree approach is used but the authors are more particularly interested in the location of the initial grid, without questioning the arbitrary choice of a square tiling of the plane.
Moreover, although the spanning tree approach is particularly efficient in terms of algorithmic cost, this point is not explored in depth in most articles.

\subsection{Traveling Salesman Problem}
An alternative casting for the coverage problem can implemented through the Travelling Salesman Problem (TSP) \cite{noauthor_travelling_2022}.The TSP belongs to the class of combinatorial optimization problems, where the objective is to find the shortest single path that, given a list of cities (or nodes) and distances between them, visits all the cities only once.

\subsection{Optimal Control Problem}
Optimal control problems present themselves in a variety of domains, such as path planning, to determine the control inputs that minimize a given cost function while fulfilling system dynamics and limitations. 

There are various types of optimal control problems, depending on the performance index, the type of time domain (continuous, discrete), the presence of different types of constraints, and what variables are free to be chosen. The formulation of an optimal control problem requires the following:
\begin{enumerate}
    \item a mathematical model of the system to be controlled,
\item a cost function,
\item a specification of all boundary conditions on states, and constraints to be satisfied by states and controls,
\item a statement of what variables are free.
\end{enumerate}    
 There are various algorithms for solving optimal control problems, such as evolutionary algorithms, shooting method approaches, and others.

Searching for some sources in an environment can be approximated as searching for uncertain targets, as expressed in \cite{walton_optimal_2014}, which has implemented the shooting approach principles. The shooting approach has been depicted in the optimal search problem, created in the 1940s, offering a fundamental framework for motion planning when looking for uncertain targets. The optimal search problem examines how to maximize the likelihood of detecting a non-evasive target with uncertain properties, given the capabilities of the detection equipment and some restrictions on searcher trajectories.

Koopman Search Theory \cite{koopman_theory_1956} was first developed during World War II to optimize anti-submarine warfare strategies. Since then, it has been extensively used in tactical and industrial search applications. This work reviews the mathematical foundation of Koopman’s search models and explores modern numerical techniques that enhance their solution methods.
	\section{Methodology}
\label{chap:Model}

All the methods discussed aim to solve the coverage problem, which consists of minimizing the path length while ensuring that the entire area is traversed efficiently. Their mathematical formulations differ, but they ultimately contribute to the same goal of optimal coverage.

\subsection{Optimal Control Problem (OCP)} \label{Optimal Control Problem:Methodology}
The problem of searching for uncertain targets can be formulated as described in \cite{walton_optimal_2014}. To construct a performance criterion for the optimal search problem, the probability of target detection must be properly modeled. According to the Koopman search theory \cite{koopman_theory_1956}, an exponential detection probability model can be derived, which has since become a fundamental framework in optimal search literature.

\subsubsection{Detection Model Implementation}
The exponential detection model assumes that the instantaneous detection rate of a target can be accurately represented. Given a searcher’s position at $p_s(t)$ and a target’s position at $p_t(t)$, this detection rate is given by $\eta(p_s(t), p_t(t))$, which determines how sensor characteristics affect detection probability. The detection process is assumed to be memoryless, meaning that detection events in different time intervals are independent.
A widely used model for defining $\eta(p_1, p_2, t)$ is the Poisson Scan Model, which describes detection as a probabilistic process influenced by distance and sensor parameters. Under this model, the detection rate function is given by:

\begin{equation}
\eta(p_s, p_t) = \lambda \Phi \left( \frac{F - a \|p_s - p_t\|^2}{\sigma} \right),
\end{equation}

where $\lambda$ is a scaling factor representing detection intensity, $\Phi(\cdot)$ is the cumulative normal distribution function, $F$ is a threshold parameter, $a$ controls the decay of detection probability with distance, and $\sigma$ is a spread parameter.

In the specific implementation used in this work, the function $\eta(p_s, p_t)$ is designed to account for spatially dependent target detection likelihoods. The detection rate function is formulated as:

\begin{equation}
\eta(p_s, p_t) = \gamma\left( \frac{\pi}{2} - \arctan \left( a\left( \frac{\| p_s - p_t \|^2}{r^2} - 1 \right) \right) \right) \frac{1}{\pi},
\end{equation}

where $\gamma$ is a scaling factor that controls the maximum detection rate, $a$ is a steepness parameter that determines how sharply the detection rate decreases with distance, $r$ is the detection radius that defines the detection sensitivity region. This function serves as an approximation of the indicator function of the circle with radius $r$, that is, when $a\rightarrow \infty$ $\eta(p_s, p_t)$ approximates

\begin{equation}
\eta(p_s, p_t) \approx \begin{cases}
\gamma & \text{if } \| p_s - p_t \| < r,\\
0  & \text{if } \| p_s - p_t \| \geq r.
\end{cases}
\end{equation}

This function defines how detection probability decays smoothly as the searcher moves further away from the target, with a controlled rate of decay dictated by the parameters above.

Under these assumptions, an explicit formula for the probability of target detection is derived. Let $P(t)$ denote the probability of non-detection at time $t$. The independence of time intervals leads to the recurrence equation:

\begin{equation}
P(t+\Delta t) = P(t) \left[1 - \eta(p_s(t), p_t(t)) \Delta t \right],
\end{equation}

where $\eta(p_s(t), p_t(t))\Delta t$ represents the probability of detection within the interval  $\Delta t$, $1 - \eta(p_s(t), p_t(t), t)\Delta t$ is the probability that the target remains undetected, and $P(t + \Delta t)$ is the updated probability of non-detection at $t + \Delta t$, obtained by multiplying $P(t)$ by the probability of non-detection in that time step.

As $\Delta t \to 0$, this recurrence relation yields the exponential solution:

\begin{equation}
P(t) = e^{-\int\limits_0^t \eta(p_1(\tau), p_2(\tau), \tau) d\tau}.
\end{equation}
A more general form is expressed as:

\begin{equation}
P(t) = G\Biggl(\int\limits_0^t \eta(p_s(\tau), p_t(\tau), \tau) d\tau \Biggr).
\end{equation}

Throughout this paper, we define $G(x)$ as:

\begin{equation}
G(x) := e^{-x}.
\end{equation}

This formulation ensures that $G(x)$ represents an exponentially decreasing probability model, aligning with the Koopman search theory.

Throughout this paper, we consider a deterministic motion model of the searchers $p_s(t)$, which assumes that the target’s motion is entirely determined by time. The potential targets are known to belong to a bounded space $\Omega \subset \mathbb{R}^n$ and their existence follow a known probability density function $p: \Omega \rightarrow \mathbb{R}$.

Given a searcher motion $p_s(t)$, the probability of not detecting a target $P(t)$ becomes a random variable. A natural performance metric is to minimize the expectation of this probability over a time interval $[0, T]$, leading to the following cost function:

\begin{equation}\label{equation:equ3.8}
    J= \int_{\Omega} G\Biggl(\int\limits_0^T \eta(p_s(t), \omega) dt \Biggr ) p(\omega) d\omega.
\end{equation}

Using the derivations in \cite{walton_optimal_2014}, Equation (\ref{equation:equ3.8}) can be discrtetized, as shown in Section \ref{subsec:OCPF}.

For the coverage problem, we consider a uniform a-priori distribution

\begin{equation}
    p(\omega) = \begin{cases}
\frac{1}{\int_\Omega d\omega} & \text{if } \omega\in\Omega,\\
0  & \text{if } \omega\notin\Omega.
\end{cases}
\end{equation}

\subsubsection{Optimal Control Problem Formulation}
\label{subsec:OCPF}

The Optimal Control Problem (OCP) aims to determine the optimal trajectory for a system while satisfying dynamic and control constraints. 

The system follows a single-integrator model where the state $p_s(t) \in \mathbb{R}^2$ evolves according to the control input $u(t) \in \mathbb{R}^2$:

\begin{equation}
\frac{dp_s}{dt} = u.
\end{equation}

This means that the control input directly determines the velocity of the system.

The system is subject to the following constraints:

\begin{align}
    &p_s(0) = p_\text{s,init}, \quad p_s(T) = p_\text{s,final}, \\
    &\| u(t) \| \leq v_\text{max}, \quad \forall t \in [0,T], \\
    &p_\text{s,min} \leq p_s(t) \leq p_\text{s,max}, \quad \forall t \in [0,T].
\end{align}

The time horizon $T$ is divided into $N$ intervals, with a timestep $\Delta t = T/N$. The discretized state and control variables are:

\begin{equation}
p_(s,k) = p_s(k \Delta t), \quad u_k = u(k \Delta t), \quad k = 0, 1, \dots, N.
\end{equation}

The system dynamics are enforced at each step using an explicit Euler integration:

\begin{equation}
p_{s,k+1} = p_{s,k} + \Delta t u_k, \quad k = 0, \dots, N-1,
\end{equation}

where control and state constraints are imposed at each discretized step.

From the probability formulation the cost function we wish to minimize is

\begin{equation}
J= \int_{\Omega} G\Biggl(\int\limits_0^T \eta(p_s(t), \omega) dt \Biggr ) p(\omega) d\omega,
\end{equation}

we approximate the inner integral with a summation:

\begin{equation}
\int_0^T \eta(p_s(t), \omega) dt \approx \sum_{k=0}^{N-1} \eta(p_{s,k}, \omega) \Delta t.
\end{equation}

Substituting this into $J$:

\begin{equation}
J \approx \sum_{i=1}^{M} G \Biggl( \sum_{k=0}^{N-1} \eta(p_{s,k}, \omega_i) \Delta t \Biggr) p(\omega_i) \Delta \omega,
\end{equation}

where $M$ is the number of evenly spaced discrete samples of $\omega$.

A penalty function is introduced to ensure the trajectory remains inside the valid search space and avoids obstacles. The penalty is computed as:

\begin{equation}
J_\text{penalty} = \sum_{k=0}^{N} \sum_{i=1}^{M} e^{-\left( \| p_{s,k} - O_i \|^2 - d^2 \right)},
\end{equation}

where $O_i$ represents the positions of obstacles, and $d$ is a threshold defining the avoidance region.

To discourage unnecessarily long paths, a regularization term penalizing the total control effort is included:

\begin{equation}
J_\text{length} = \lambda \sum_{k=0}^{N-1} \| u_k \| \Delta t.
\end{equation}

This acts as a soft constraint, pushing the optimization towards minimizing unnecessary deviations.

The overall objective function minimized in the OCP is:

\begin{equation}
J = \sum_{i=1}^{M} G \Biggl( \sum_{k=0}^{N-1} \eta(p_{s,k}, \omega_i) \Delta t \Biggr) p(\omega_i) \Delta \omega + J_\text{penalty} + J_\text{length},
\end{equation}

where the first term represents the primary cost term associated with optimizing search efficiency, and the additional terms enforce obstacle avoidance and path regularization.

 This results in a computationally efficient numerical approach to solving the OCP using direct multiple-shooting methods in MATLAB. To improve computational efficiency and provide a structured initial guess for the Optimal Control Problem (OCP), the solution obtained from the TSP heuristic is used as an initial trajectory input for the solver. The optimization problem is then solved using nonlinear programming techniques, ensuring an optimal trajectory while satisfying system constraints. By leveraging the TSP path, the OCP solver benefits from a well-structured initial condition, leading to faster convergence and improved feasibility of solutions.

\subsection{Minimum Spanning Tree }\label{MST_explanation}

We can distinguish several specificities specific to the marine environment:
\begin{enumerate}
\item the presence of a third dimension
\item winds and sea currents fluctuating over time
\item the difficulty of navigating certain robots (sailboats, gliders, etc.)
\item the risk of losing a robot (risks of collision, breakdowns, etc.)
\item the specificity of sensors in a marine environment
\item the difficulty of communicating with fixed stations or between robots when they are underwater
\end{enumerate}
We will not go into details in this article because our study is a preliminary work. However, we keep in mind certain objectives in order to overcome the difficulties stated below. Thus, we must generate admissible trajectories, favoring straight lines. But above all, we want to be able to solve the problem very quickly with very limited computing resources, so as to recalculate solutions with on-board computing and in real time, so as to be able to react to disturbances (winds, currents, anti-collision diversion)

We will reformulate the problem as an optimal coverage problem, for which we can find a straightforward solution. Rather than maximizing a criterion representing the probability of discovering a source—which intuitively corresponds to meticulously scanning the search area while covering every part—we will explicitly define the problem as finding a path that ensures no point remains at a distance greater than a given threshold, which depends on the sensor’s range. We denote this distance as $r$. 
Let us remark that if the probability of detection of the sensor is equal to 1 inside the disk of radius r centered on the sensor and 0 outside the disk, minimizing the probability of non-detection is exactly the same as our reformulated problem.

Thus, the goal is to find a closed path (so that the robot returns to its starting position) that explores a given area in such a way that every point within this area is covered at some moment by the sensor. This is exactly the problem encountered in robotic vacuum cleaners.
This type of problem is relatively easy to solve in the case of a rectangular area. To simplify the description, we will assume that the rectangle’s corners are rounded with quarter-circle arcs of radius $r$ (since a robotic vacuum cleaner cannot clean sharp corners, but in our case, this assumption serves merely as a convenient way to explain the algorithm). We will also assume that the lengths of the rectangle’s sides are multiples of $2r$. Again, this assumption simplifies the exposition of the method and helps convince the reader of the method’s near-optimality. Indeed, in this case, the well-known optimal solution is obtained by sweeping the area using a zigzag trajectory. For a glider-type robot, such a trajectory is particularly interesting because it favors straight-line motion (we will disregard the curvature of the Earth).

 \begin{figure}[H]\centering
\captionsetup{justification=centering}
	\begin{tikzpicture} 
	\draw[draw=none,fill=gray!50] (0,0.5) rectangle (3,1.5);
	\draw (0,0.5) -- (3,0.5);
	\draw (0,1.5) -- (3,1.5);
	\draw[dotted, thick] (0,1) node[left] {\scriptsize spanning tree line} -- (3,1);
	 \draw [decorate,decoration={brace,mirror,amplitude=3}]  (3.1,0.5) -- (3.1,1.5) node [midway,right,xshift=.1cm] {\scriptsize dilated line};
	\draw[dashed] (0,2) -- (3,2);
	\draw[fill=none] (1,1.5) circle (0.5);
	\draw [->,>=stealth,thick] (1.5,1.5) -- (2,1.5);
	\draw[fill=none] (2,0.5) circle (0.5);
	\draw [->,>=stealth,thick] (1.5,0.5) -- (1,0.5);
	\draw[dashed] (0,0) -- (3,0);
	\end{tikzpicture}
\caption{Main principle for the covering\label{fig:construction}}
\end{figure}
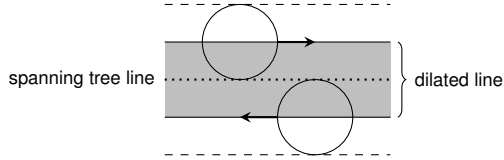

The area to be explored is not necessarily rectangular and may not be 1-simply connected. Therefore, the zigzag trajectory must be adapted according to the boundaries and the regions to be avoided. The challenge is thus to develop an algorithm that naturally generates this type of trajectory within a rectangular domain but can also extend to more complex cases.

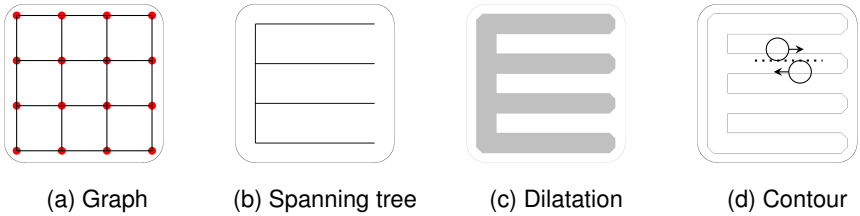
\begin{figure}[H]\centering
\captionsetup{justification=centering}
\begin{subfigure}{0.24\textwidth}
	\begin{tikzpicture}
	\node at (0,0) {\includegraphics[width=.8\textwidth]{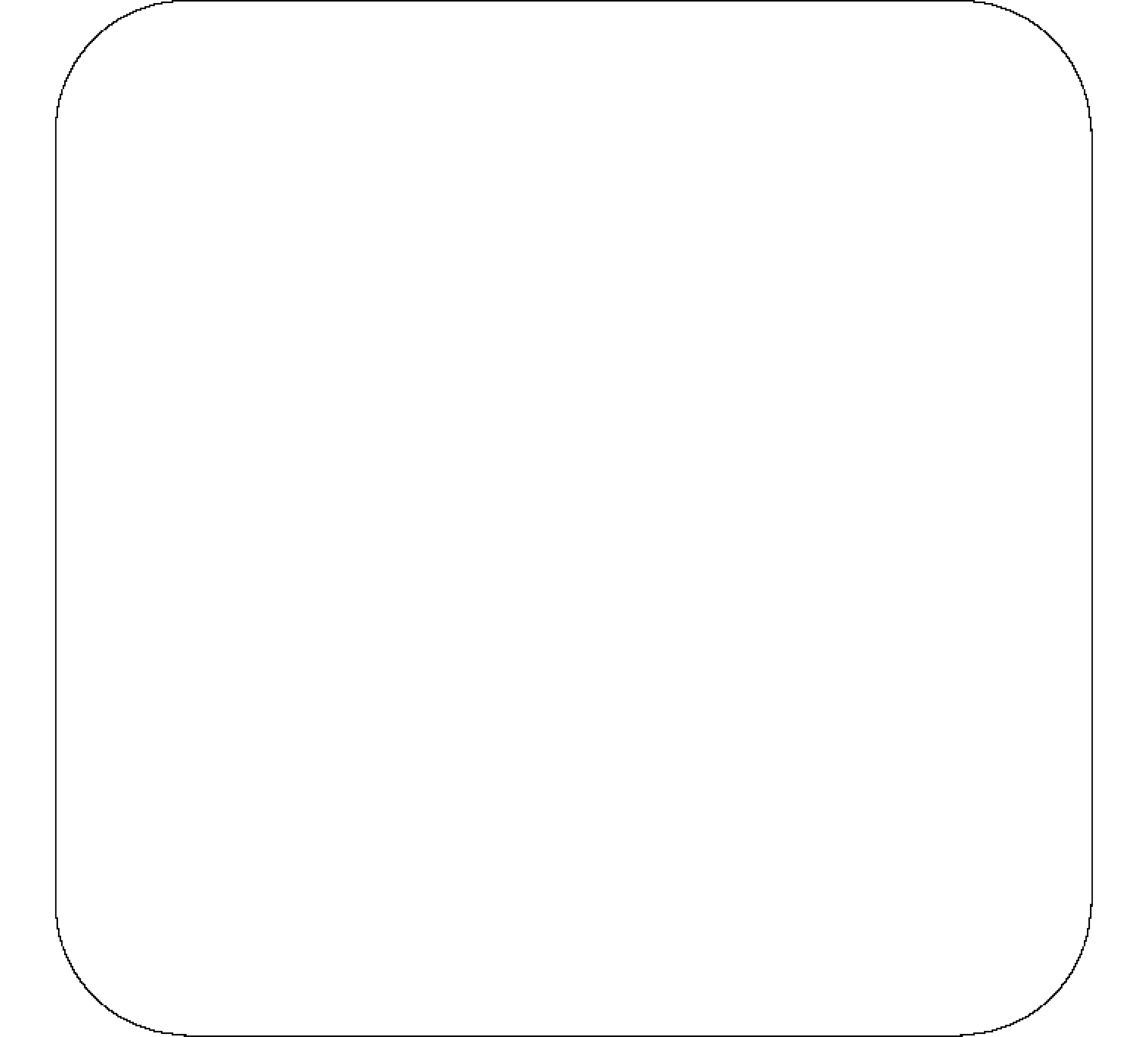}};
	\foreach \x in {-0.9,-0.3, ..., 0.9} {
	    \foreach \y in {-0.9, -0.3, ..., 0.9} { 
                \fill[red] (\x,\y) circle[radius=1.5pt];        }};
	\draw[step=0.6,black,xshift=-0.9cm,yshift=-0.9cm] (0,0) grid (1.8,1.8);
	\end{tikzpicture}
\caption{Graph\label{fig:20:graph}}
\end{subfigure}
\hfill
\begin{subfigure}{0.24\textwidth}
	\begin{tikzpicture}
	\node at (0,0) {    \includegraphics[width=.8\textwidth]{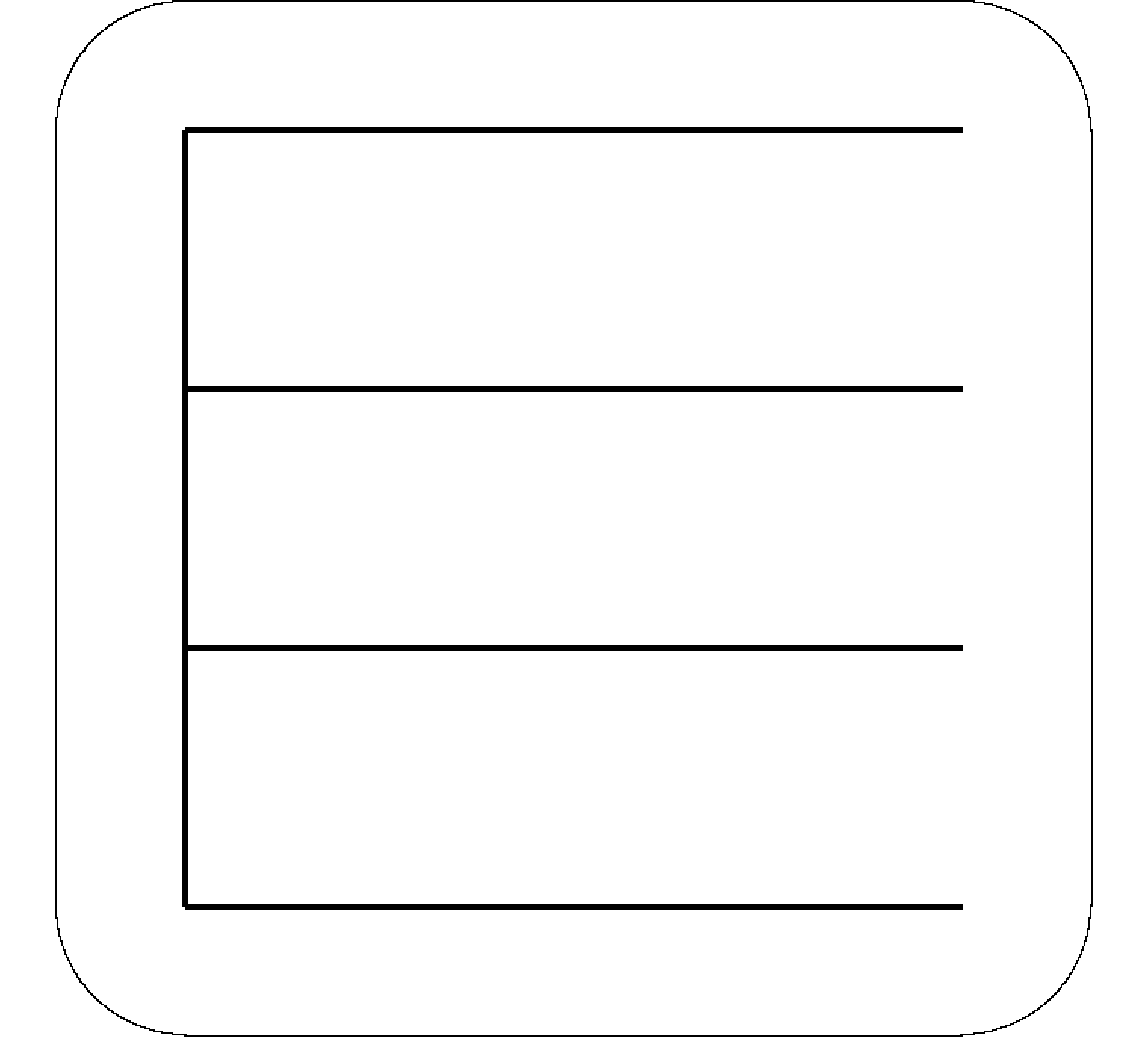}};
	\end{tikzpicture}
\caption{Spanning tree \label{fig:20:MST}}
\end{subfigure}
\hfill
\begin{subfigure}{0.24\textwidth}
	\begin{tikzpicture}
	\node at (0,0) {    \includegraphics[width=.8\textwidth]{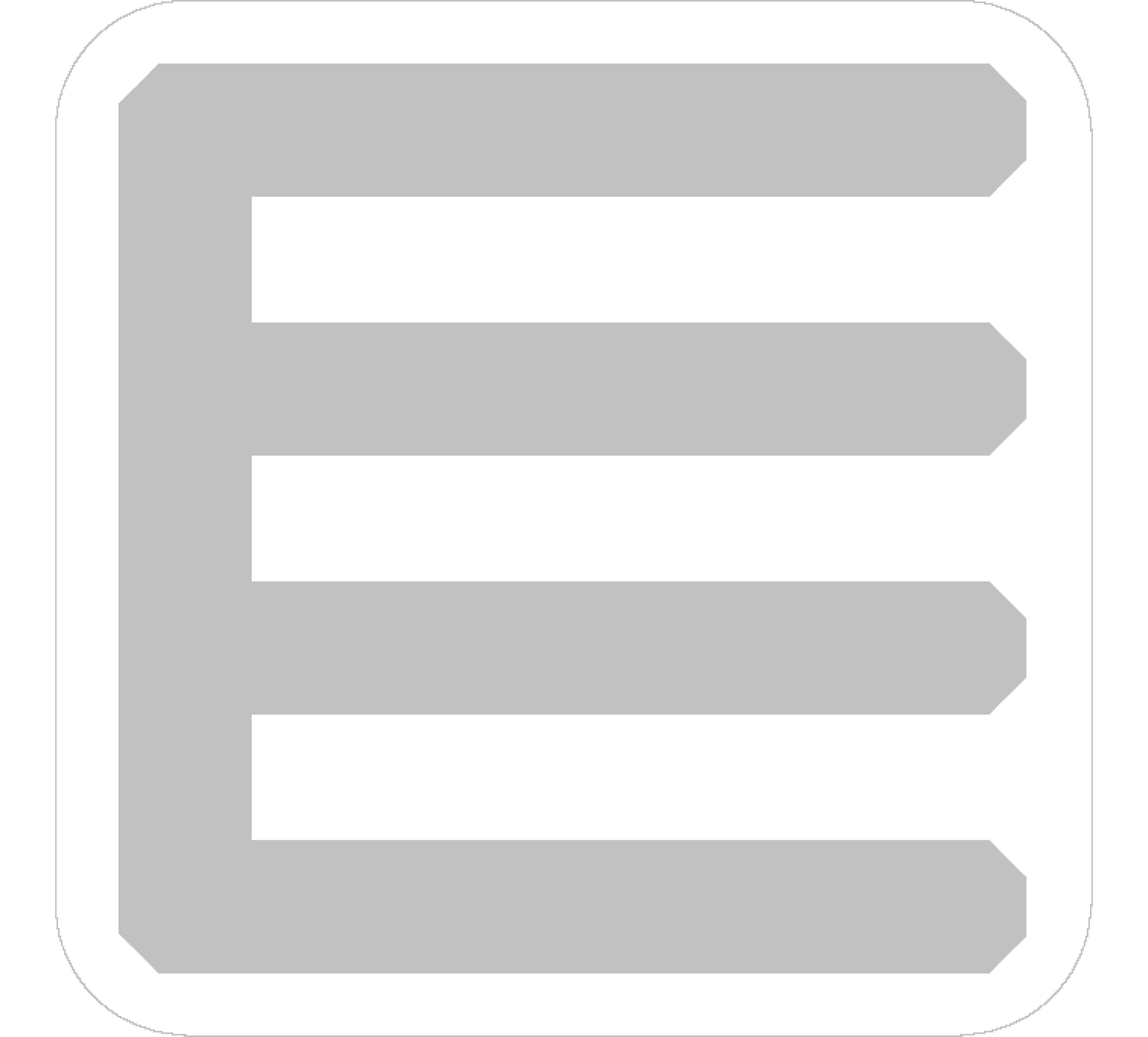}};
	\end{tikzpicture}
\caption{Dilatation \label{fig:20:dilatation}}
\end{subfigure}
\hfill
\begin{subfigure}{0.24\textwidth}
	\begin{tikzpicture}
	\node at (0,0) {\includegraphics[width=.8\textwidth]{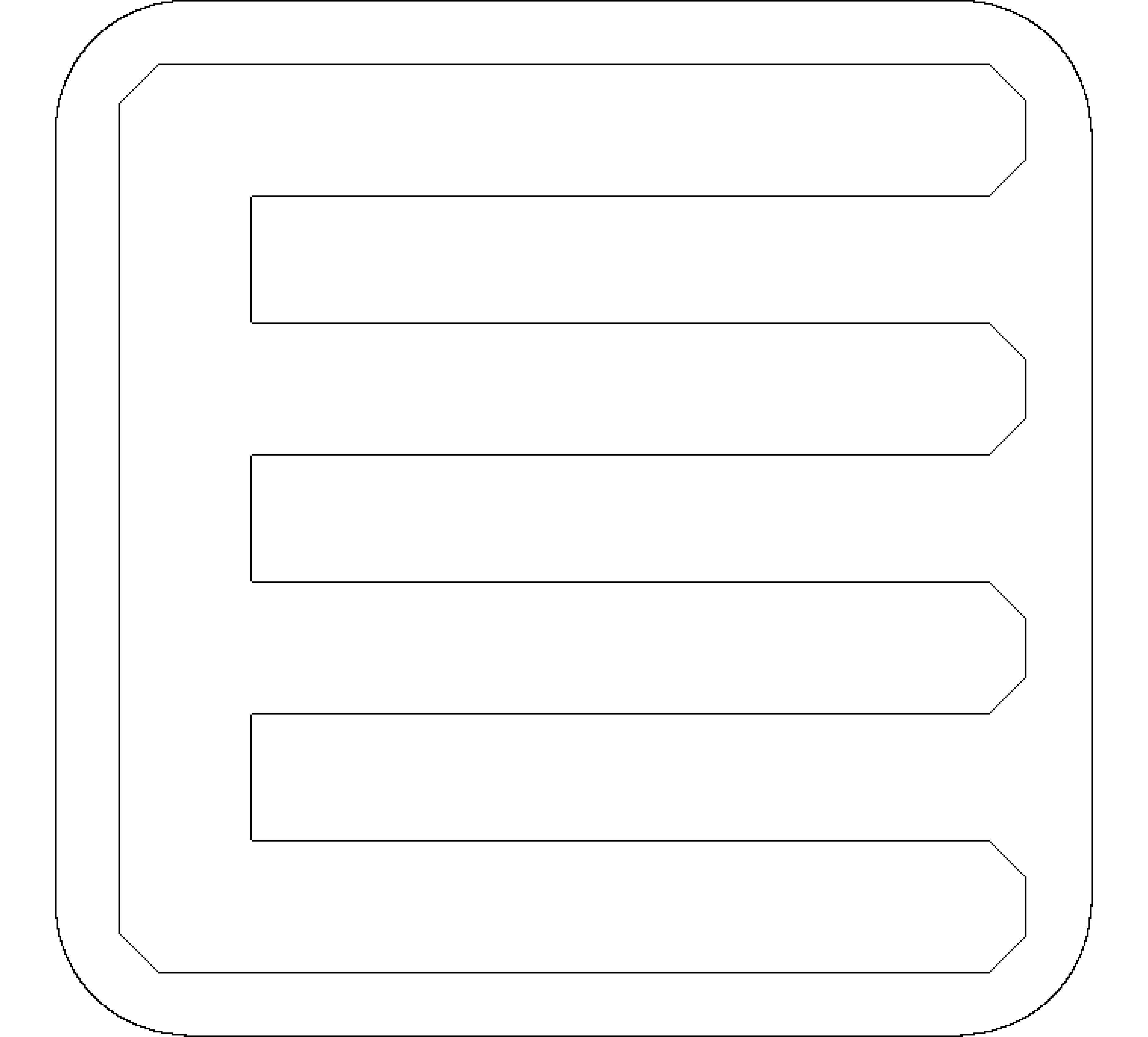}};
	\draw[fill=none] (0.0,0.45) circle (0.15);
	\draw [->,>=stealth] (0.15,0.45) -- (0.35,0.45);
	\draw[dotted,thick] (-0.3,0.3) -- (0.6,0.3);    
	\draw[fill=none] (0.3,0.15) circle (0.15);
	\draw [->,>=stealth] (0.15,0.15) -- (-0.05,0.15);
	\end{tikzpicture}
\caption{Contour \label{fig:20:contour}}
\end{subfigure}
\caption{Construction of trajectory : (a), we compute a minimal spanning tree (b), we perform a dilatation of this spanning tree of radius $r$ (c) and we finally obtain a trajectory as the border of the dilated tree\label{fig:20}}
\end{figure}

To achieve this,  the surface to be explored is meshed using a $2r$-grid which constitutes a graph $\mathcal G$ whose intersection points form the vertices, see Figure \ref{fig:20:graph}. Two vertices are connected if they are adjacent in the 4-neighborhood sense. The algorithm, which is straightforward, is as follows:  

We first extract a minimum spanning tree (MST) from this graph, see Figure \ref{fig:20:MST}. From a graph-theoretical perspective, we traverse this tree using a depth-first search (DFS). Geometrically, this corresponds to traversing the tree while maintaining a distance $r$ from it. To visualize (and compute) this trajectory, we simply apply a morphological dilation to this graph—interpreted as an image processing operation—with a structuring element of radius $r$, see Figure \ref{fig:20:dilatation}. Finally, we extract the contour of the dilated result, see Figure \ref{fig:20:contour}. 
This contour defines the trajectory to be followed. By construction, it covers the whole space. 

The total computational complexity of this algorithm is approximately $\mathcal{O}(\left({\frac Sr}\right)^2 \log \left({\frac Sr}\right))$, where $S$ is the surface area of the exploration domain:  

\begin{itemize}
\item The extraction of a minimum spanning tree is performed using Kruskal’s algorithm, a greedy algorithm that can be proven to be optimal. It proceeds by sorting the edges in ascending order and iteratively adding them to the tree, ensuring no cycles are formed. Since the number of vertices is proportional to the area to be explored, the number of edges is proportional to the square of this area.  
\item The dilation operation is equivalent to a convolution with a disk of radius $r$. Assuming $r$ is small relative to the exploration area, the cost of this operation is linear with respect to the surface area.  
\item Contour extraction can also be formulated as a convolution, for instance, using the Sobel edge detection method, which remains a linear-time operation in terms of surface area. However, to obtain the final trajectory, it is more practical to perform contour following, which can also be executed in linear time.  
\end{itemize}

\begin{figure}[ht]
  \centering\captionsetup{justification=centering} 
\captionsetup{justification=centering}
\begin{subfigure}[b]{.49\textwidth}
    \includegraphics[width=\textwidth]{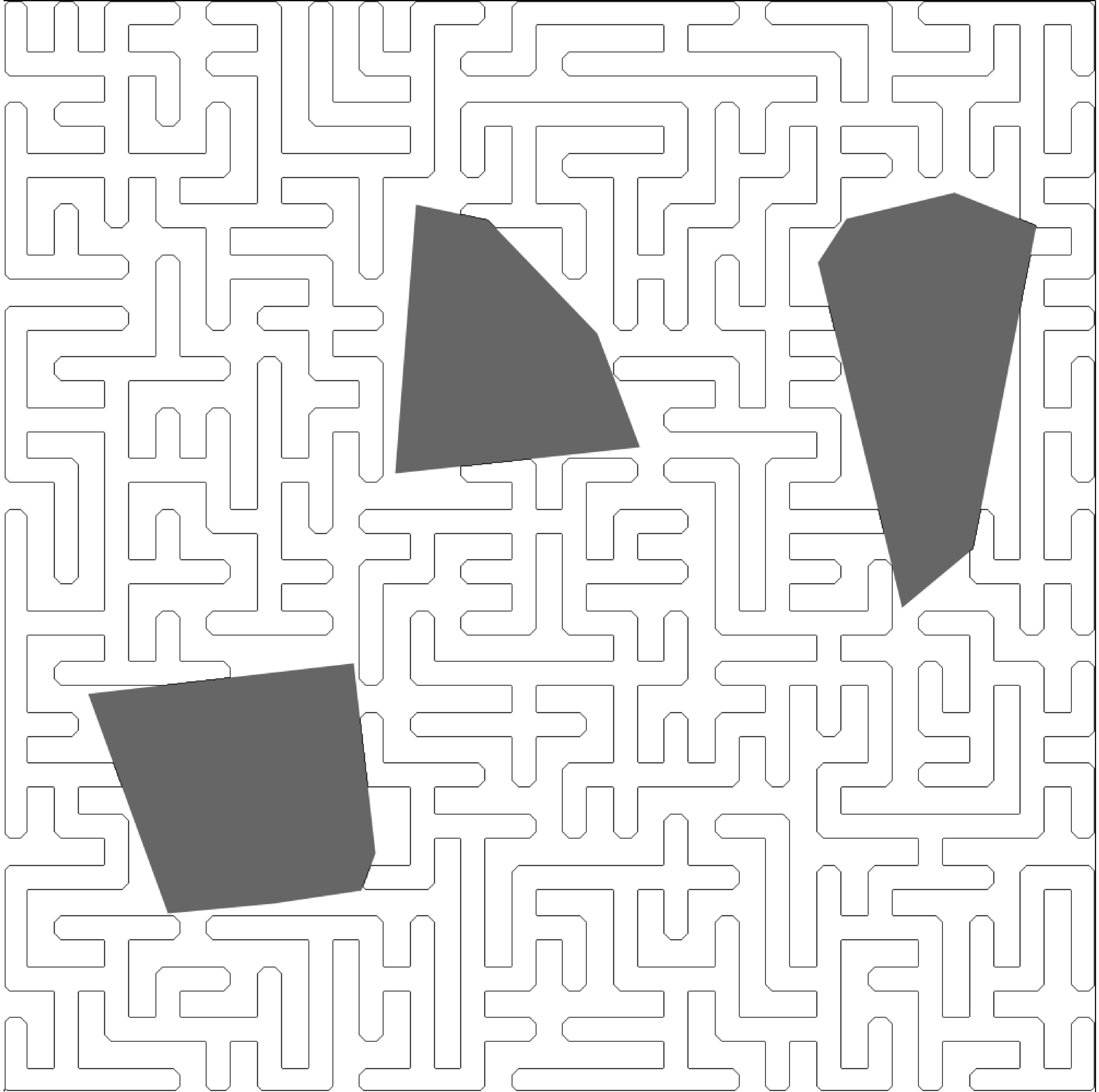}
\caption{Square grid\label{fig:euclid:sqr}}
\end{subfigure}
\hfill
\begin{subfigure}[b]{.49\textwidth}
    \includegraphics[width=\textwidth]{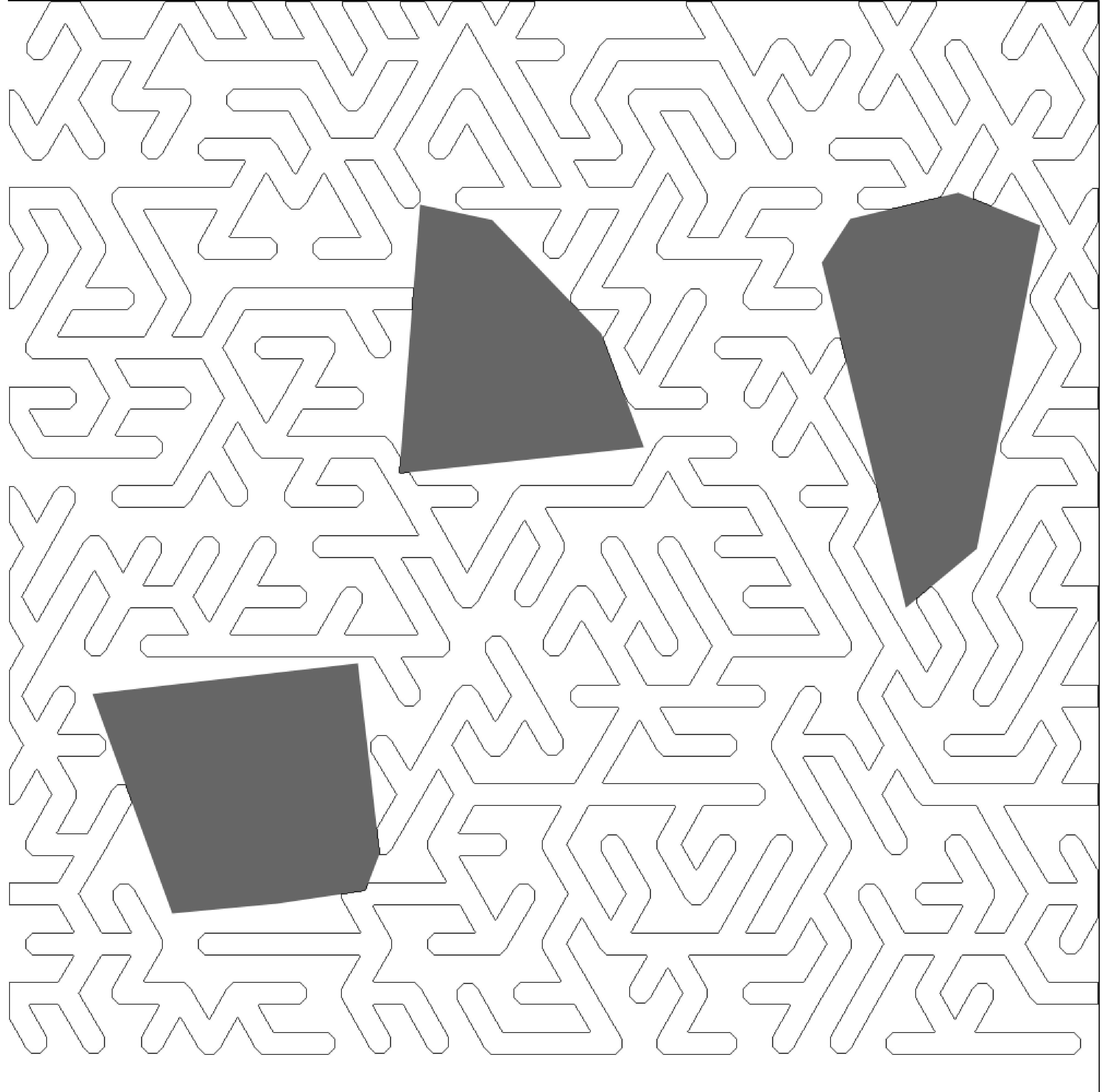}
\caption{Hexagonal grid\label{fig:euclid:hex}}
\end{subfigure}
\caption{Same algorithm with Euclidian distance\label{fig:euclid}}
\end{figure}

It is worth noting that, throughout this work, we have chosen to favor horizontal over vertical trajectories (in the orientation of our illustrations, this corresponds to east-west rather than north-south on a map—an entirely arbitrary choice). This preference is enforced by introducing a slight penalty on vertical edges compared to horizontal ones before extracting the minimum spanning tree, thereby biasing the solution toward horizontal paths. Geometrically (which becomes relevant if we later opt for a non-grid-based discretization), this corresponds to modifying the Euclidean distance into an anisotropic metric of the form  
\begin{equation}
d = \sqrt{x^2 + \varepsilon\,y^2}\label{eq:dist}
\end{equation}
where $\varepsilon$ is a small penalty parameter. The Euclidean distance provides a less convincing trajectory, see Figure \ref{fig:euclid}. 


The path we compute corresponds to a depth-first traversal of a tree extracted from this graph. Several remarks can be made at this stage:  

\begin{itemize}
\item The choice of a grid-based discretization using a rectangular tiling is arbitrary. We have also tested a hexagonal tiling (with centers still at a minimum distance of \( 2r \)), which may offer advantages for certain spatial geometries in the exploration domain, see Figures \ref{fig:hex:gridST} to \ref{fig:hex:result};  

\begin{figure}[ht]\centering
\captionsetup{justification=centering}
\begin{subfigure}[b]{.49\textwidth}\centering
    \includegraphics[width=\textwidth]{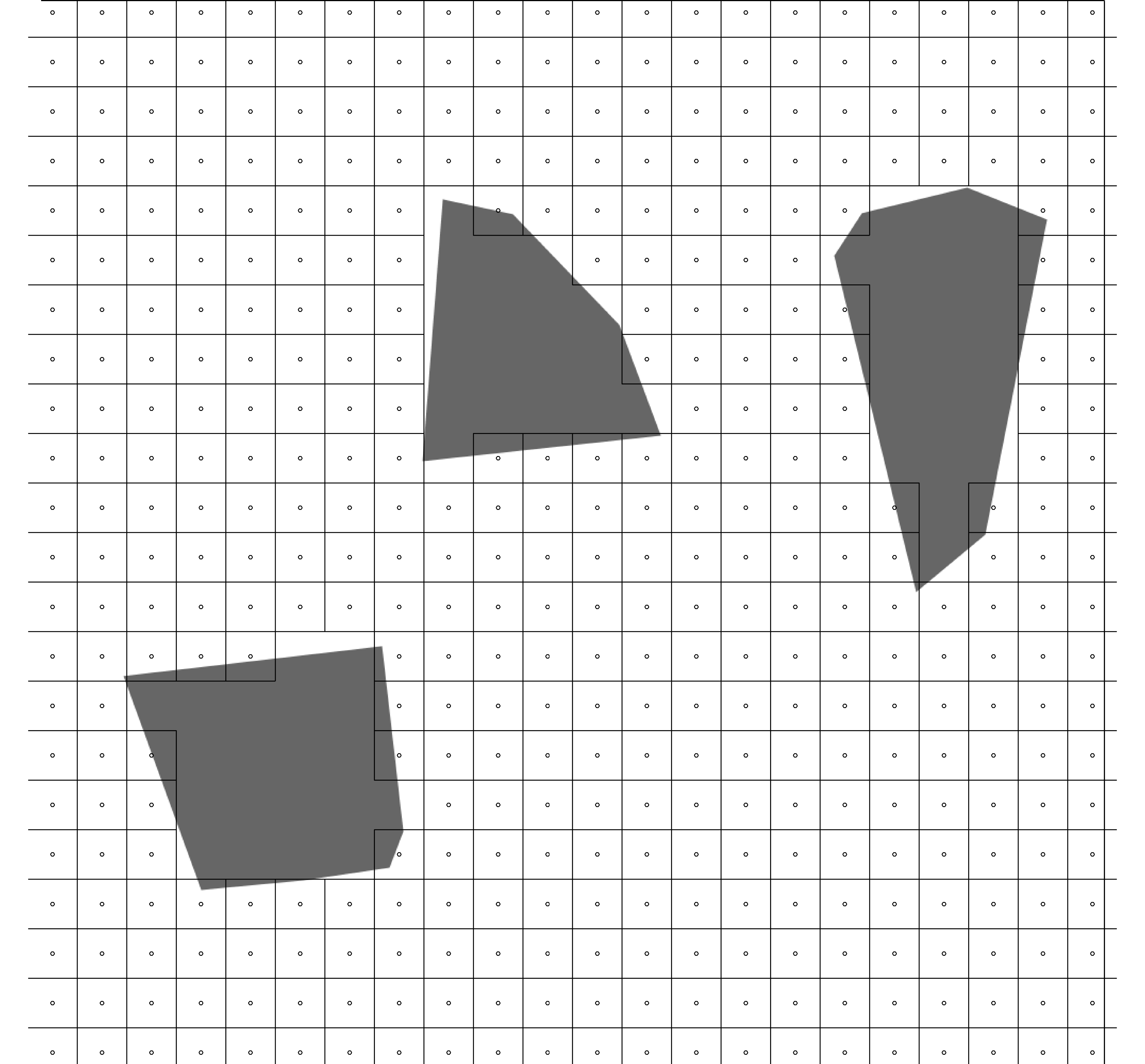}
    \caption{The grid}
    \label{fig:hex:grid}
\end{subfigure}
\hfill
\begin{subfigure}[b]{.49\textwidth}\centering
    \includegraphics[width=\textwidth]{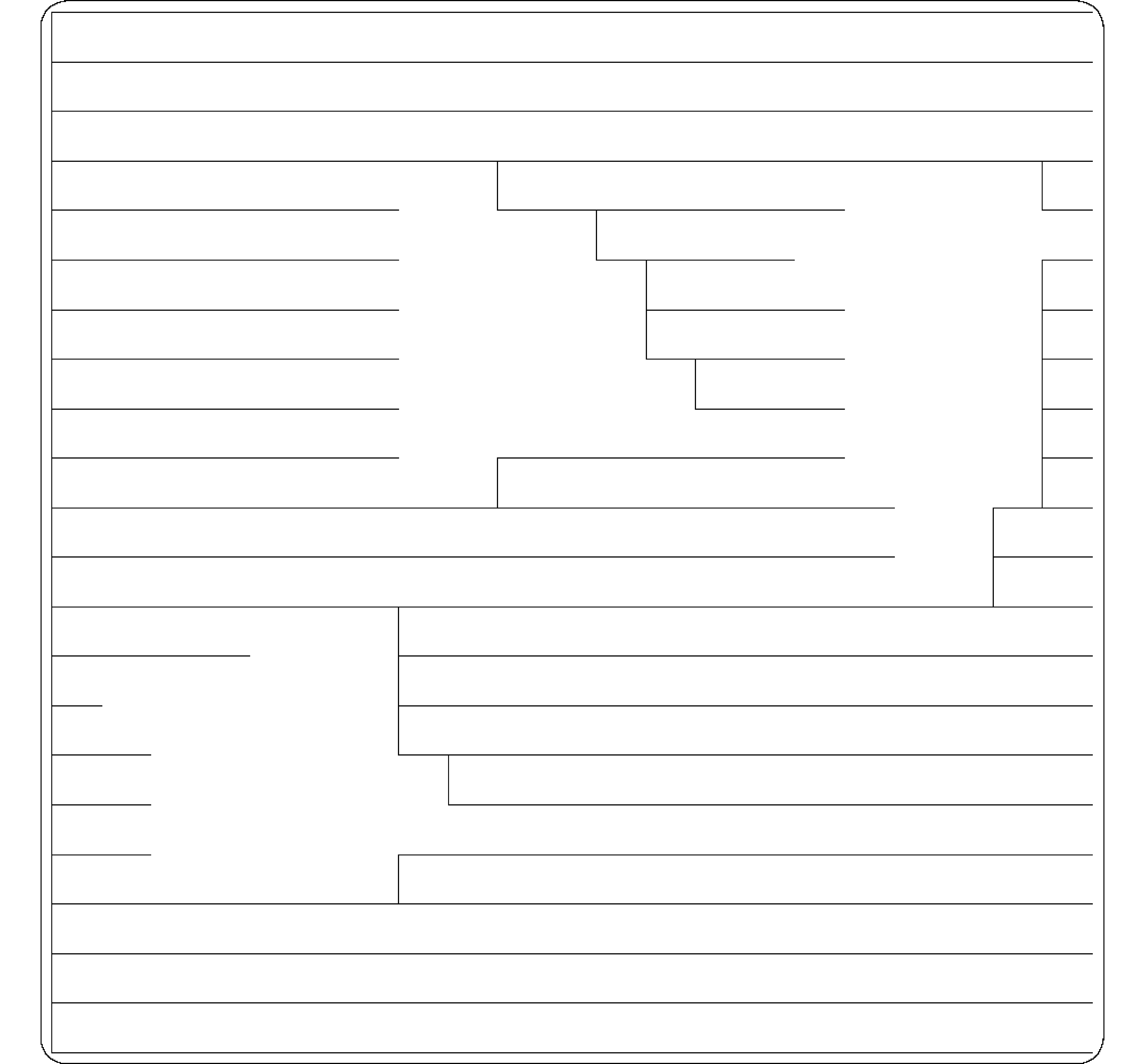}
    \caption{Spanning tree}
    \label{fig:hex:ST}
\end{subfigure}
\caption{Square grid\label{fig:hex:gridST}}
\end{figure}

\begin{figure}[ht]\centering
\captionsetup{justification=centering}
\begin{subfigure}[b]{.49\textwidth}\centering
    \includegraphics[width=\textwidth]{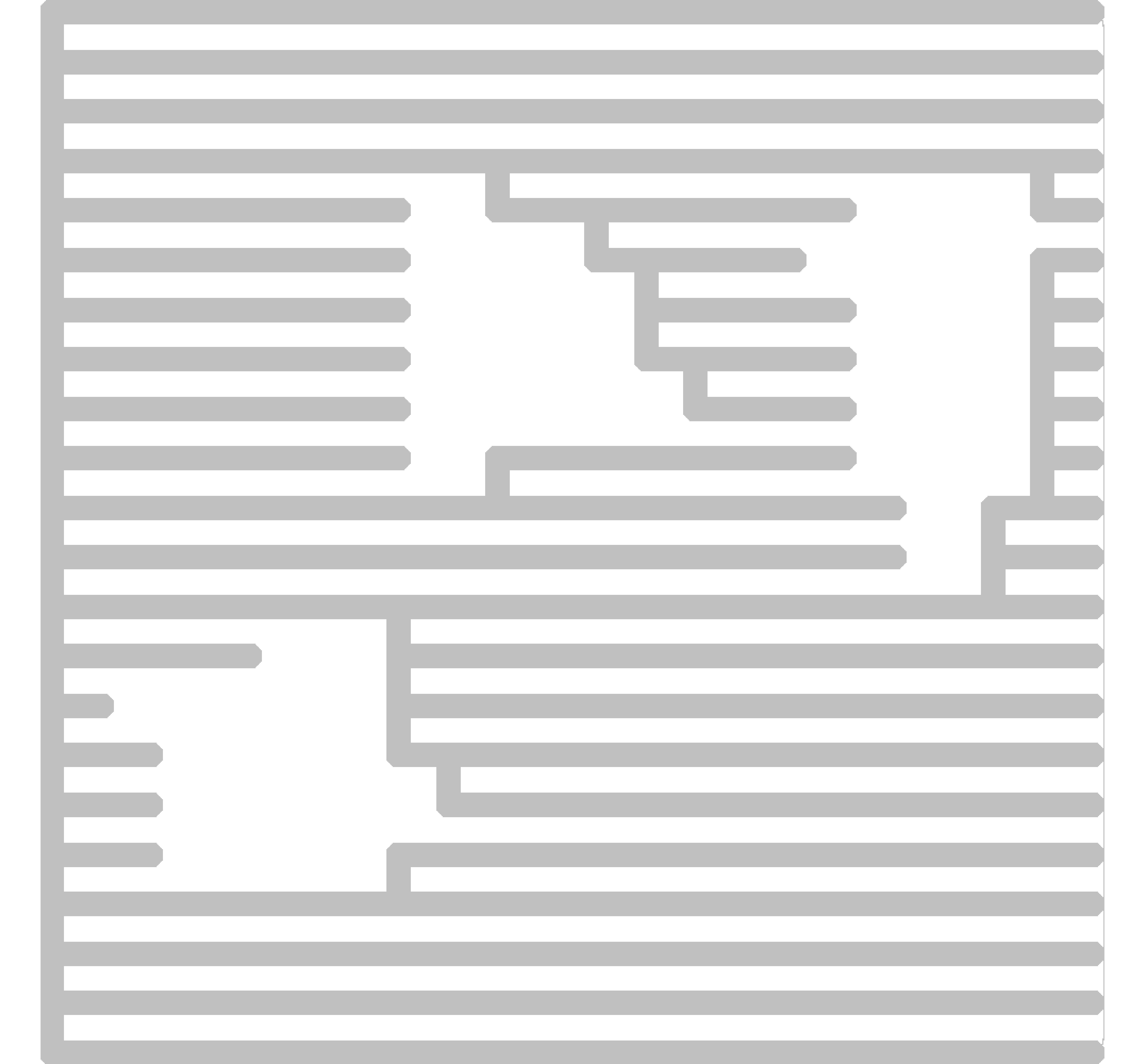}
    \caption{Dilatation}
    \label{fig:hex:dilataion}
\end{subfigure}
\hfill
\begin{subfigure}[b]{.49\textwidth}\centering
    \includegraphics[width=\textwidth]{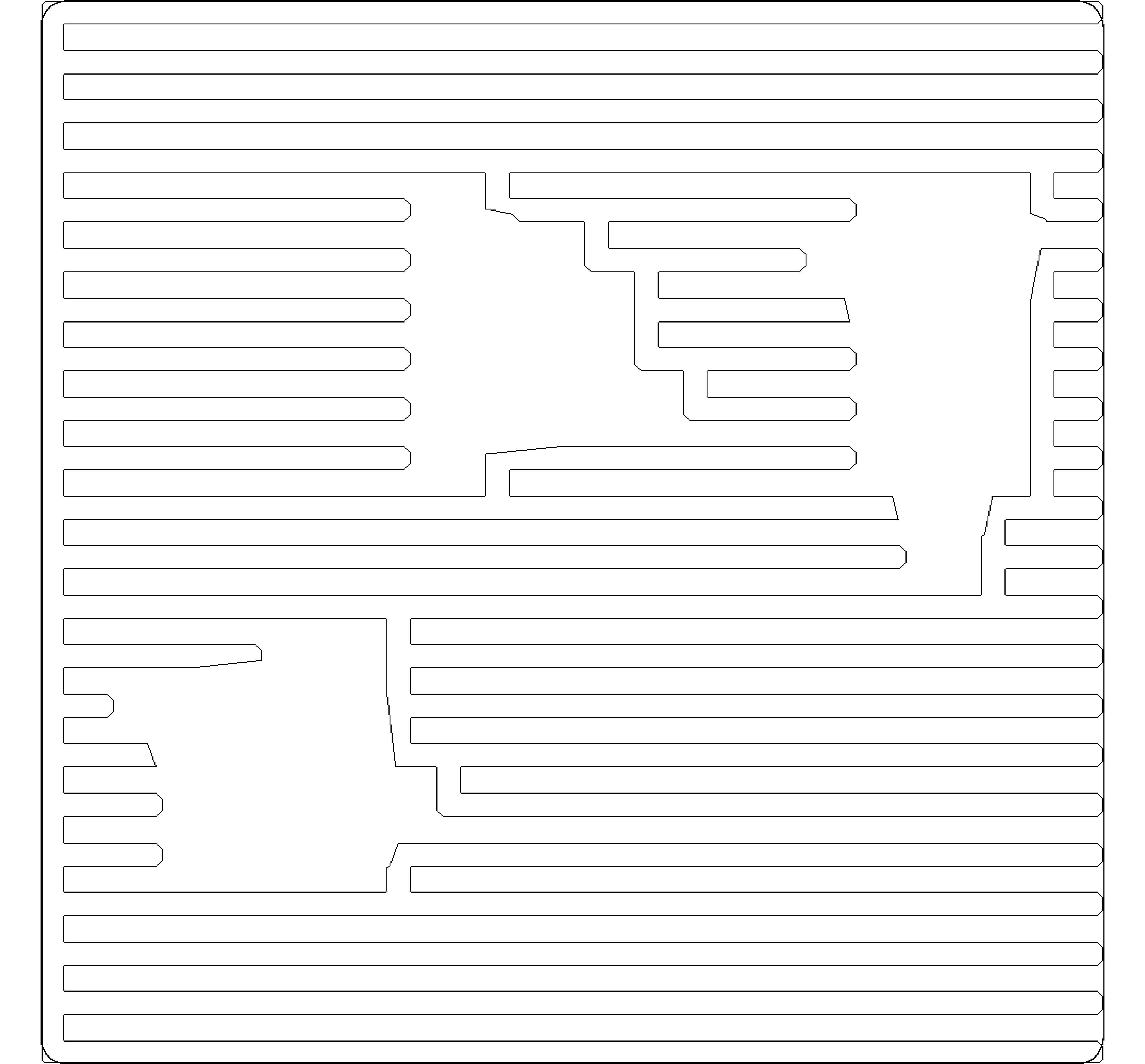}
    \caption{Contour}
    \label{fig:hex:contour}
\end{subfigure}
\caption{Algorithm\label{fig:hex:algo}}
\end{figure}

\begin{figure}[ht]\centering
\captionsetup{justification=centering}
\begin{subfigure}[b]{.49\textwidth}\centering
    \includegraphics[width=\textwidth]{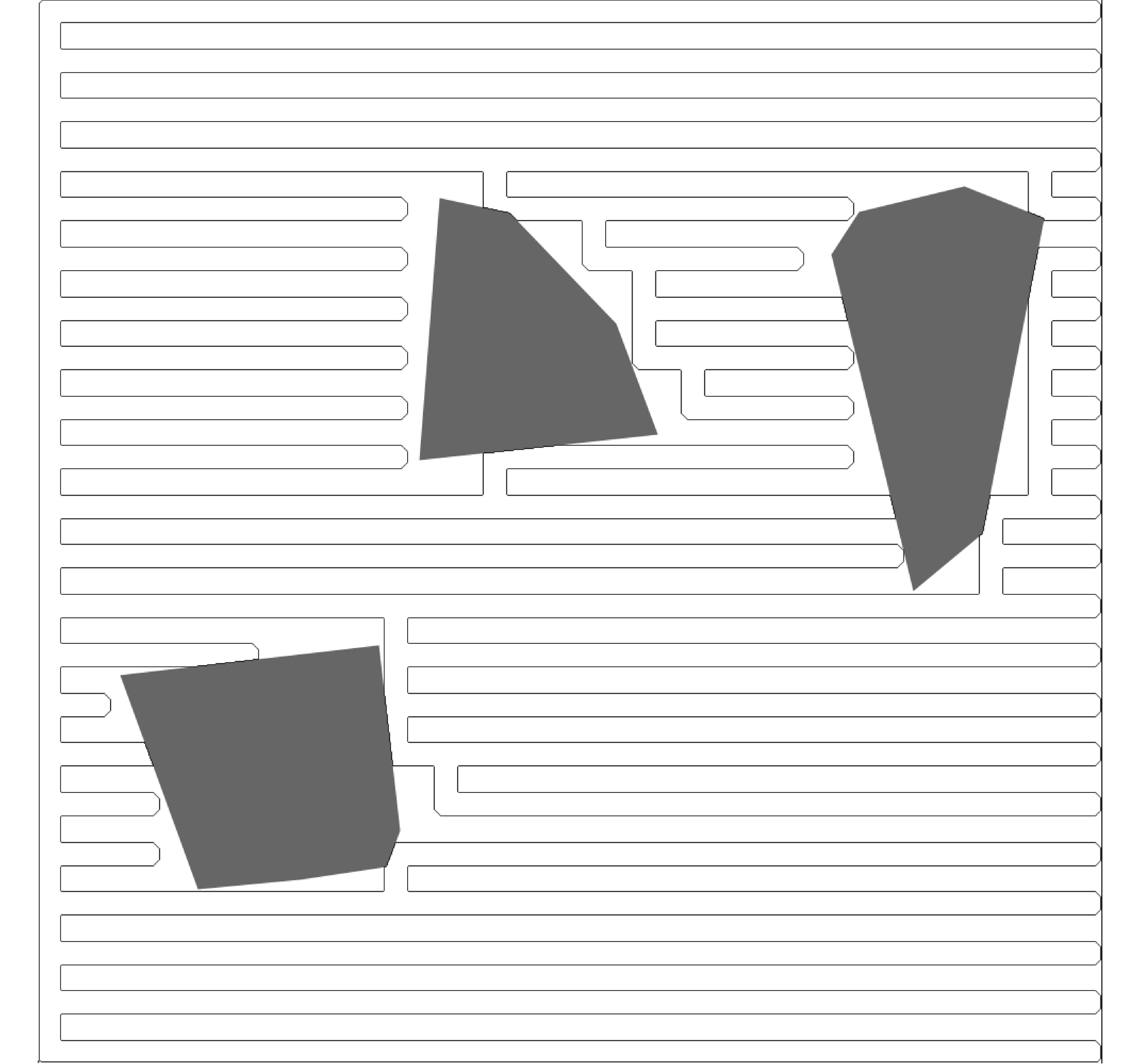}
    \caption{Final path}
    \label{fig:hex:path}
\end{subfigure}
\hfill
\begin{subfigure}[b]{.49\textwidth}\centering
    \includegraphics[width=\textwidth]{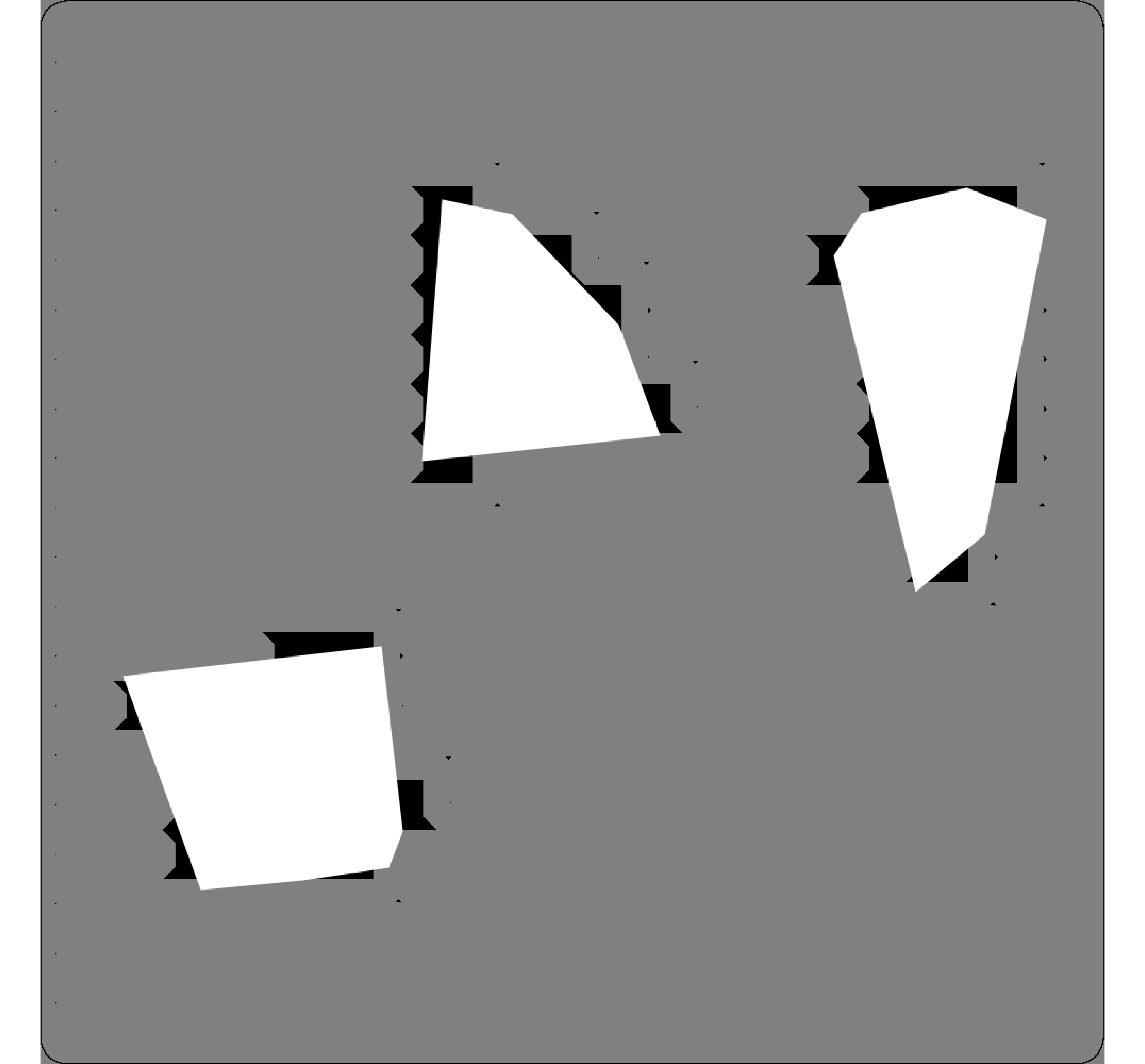}
    \caption{Uncovered area}
    \label{fig:hex:ua}
\end{subfigure}
\caption{Construction of the path\label{fig:hex:result}}
\end{figure}

\item The use of a regular grid ensures minimal-length trajectories in a rectangular domain. However, it may be beneficial to introduce additional points in the discretization to refine exploration in specific areas or to navigate through narrow passages.  
\end{itemize}

\subsection{Traveling Salesman Problem } 

The Traveling Salesman Problem (TSP) \cite{noauthor_travelling_2022} is a well-known combinatorial optimization problem. It involves finding the shortest possible tour that visits each city exactly once and returns to the starting point. TSP can be formulated as an Integer Linear Programming (ILP) problem, where binary decision variables determine whether an edge (i.e., a connection between two cities) is included in the tour.

\subsection{Problem Statement}

Consider a set of $n$ cities $C = \{1, 2, \ldots, n\}$, where $d_{ij}$ represents the distance between cities $i$ and $j$. The objective of TSP is to minimize the total tour length $L$ for a given permutation of cities. In the MATLAB implementation, instead of predefined cities, nodes are sampled on a regular grid, and obstacles are incorporated by removing points that lie within them.

\subsection{Objective Function}

The total tour length $L$ is expressed as the sum of the distances traveled between consecutive cities in the tour:

\begin{equation}
    L = \sum_{i=1}^{n} \sum_{j \neq i} d_{ij} x_{ij}.
\end{equation}

\subsection{Constraints}

To ensure a valid tour, the following constraints are imposed:

\begin{align}
    &\sum_{j \neq i} x_{ij} = 1, \quad \forall i \in C \label{eq:visit_once} \\
    &\sum_{i \in C} x_{ij} = 1, \quad \forall j \in C \label{eq:leave_once}\\
    & \sum_{i\in S} \sum_{j\notin S} x_{ij} \geq 1, \quad \forall S \subseteq \{1, 2, \ldots, n\}, S \neq \emptyset, S \neq C \label{eq:subtour_cycle}
\end{align}

Equation \eqref{eq:visit_once} ensures that each city is visited exactly once, while Equation \eqref{eq:leave_once} ensures that the salesman leaves each city exactly once. Equation \eqref{eq:subtour_cycle} prevents subtours that do not include all cities.

Similar to the Minimum Spanning Tree (MST), vertices are selected at a specified distance; however, in TSP, the distance is set to $0.5d$ instead of $d$. The MATLAB implementation constructs a graph representation of TSP rather than explicitly solving an ILP. Nodes are sampled on a mesh grid, and edges that intersect obstacles are penalized with a large distance to prevent infeasible routes.

\subsection{Implementation Considerations}

In the MATLAB implementation, obstacles are incorporated by removing grid points that lie within them. The edges of the graph are filtered to eliminate paths that intersect obstacles by assigning them a high penalty distance. Instead of an ILP formulation, a distance matrix is constructed, and the TSP solution is found using a heuristic search algorithm.

The solver applies a nearest neighbor heuristic to construct an initial tour, followed by 2-opt heuristics to improve the solution. The starting points for the nearest neighbor tour are randomly selected to explore different initial conditions. The 2-opt method iteratively swaps pairs of edges to reduce the total tour length, ensuring a more optimized route.

The objective function in TSP seeks to minimize the total distance traveled while ensuring each node is visited exactly once and the tour returns to the starting point:

\begin{equation}
    \text{Minimize} \quad \sum_{i=1}^{N} \sum_{j=1}^{N} c_{ij} \cdot x_{ij}
\end{equation}

where $N$ represents the number of nodes, $c_{ij}$ is the distance or cost between nodes $i$ and $j$, and $x_{ij}$ is a binary decision variable indicating whether travel occurs between nodes $i$ and $j$. In practice, the MATLAB implementation avoids an explicit ILP solution and instead constructs a feasible graph representation, solving for an approximate shortest path using heuristics. The heuristic approach follows a well-established method combining nearest neighbor initialization with 2-opt optimization. This approach is widely recognized in TSP research, particularly in heuristic solvers \cite{lin1965computer}.

	\section{Numerical Results}
\label{chap:Results}

This section presents results from a single reference map (for illustrative trajectories) and summarizes the average performance of four algorithms (TSP, MST-hex, MST-square, and OCP) over 10 randomly generated samples. Each algorithm aims to cover the environment efficiently, subject to obstacle constraints and sensor detection capabilities. The code used to obtain the results can be found in \href{ https://github.com/ffcrego87/cover\_planning}{github.com/ffcrego87/cover\_planning} \cite{rego2025coverplanning}.

\subsection{Single-Sample Illustration}

\begin{figure}[H]
  \centering
  \includegraphics[width=\textwidth]{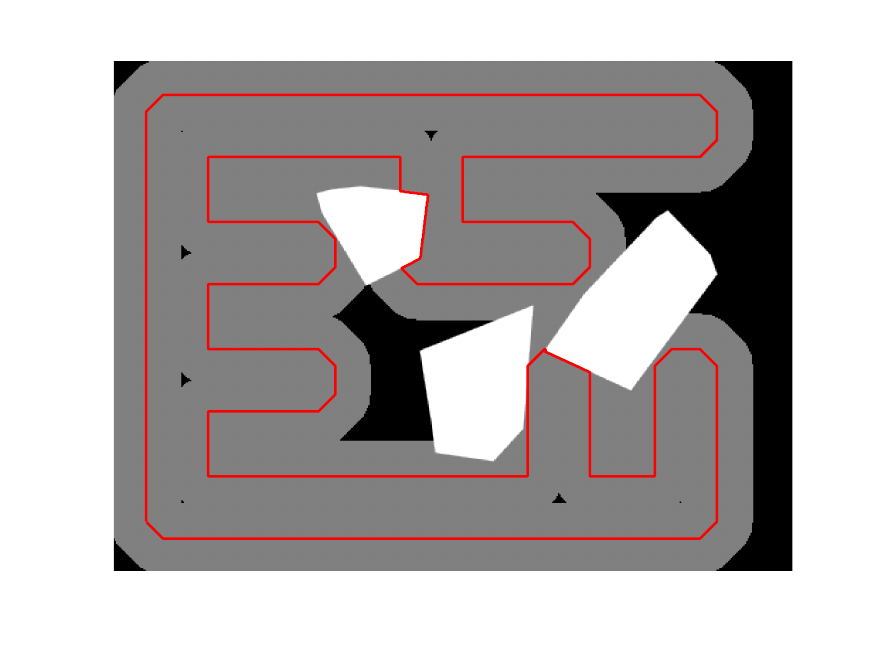}
  \caption{Representative trajectory: MST-squared path on a single map.}
  \label{fig:MSTsquareExample}
\end{figure}

\begin{figure}[H]
  \centering
  \includegraphics[width=\textwidth]{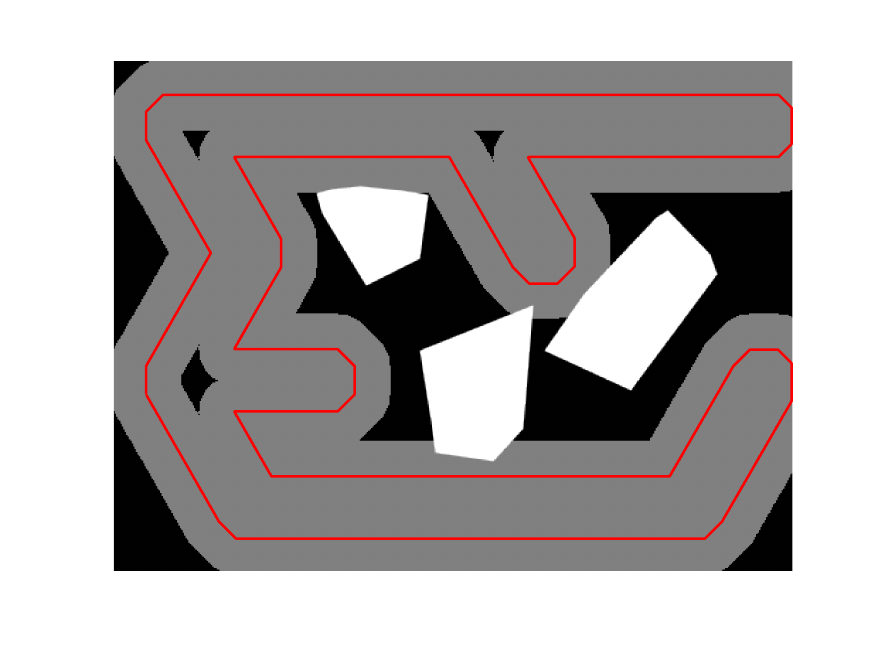}
  \caption{Representative trajectory: MST-hex path on a single map.}
  \label{fig:MSThexExample}
\end{figure}

\begin{figure}[H]
  \centering
  \includegraphics[width=\textwidth]{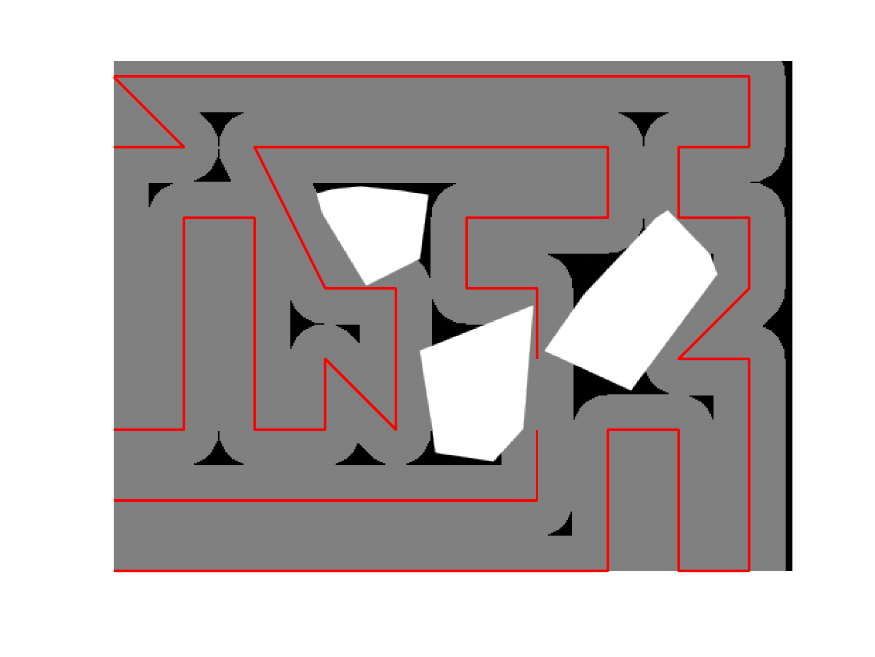}
  \caption{Representative trajectory: TSP path on a single map.}
  \label{fig:TSPExample}
\end{figure}

\begin{figure}[H]
  \centering
  \includegraphics[width=\textwidth]{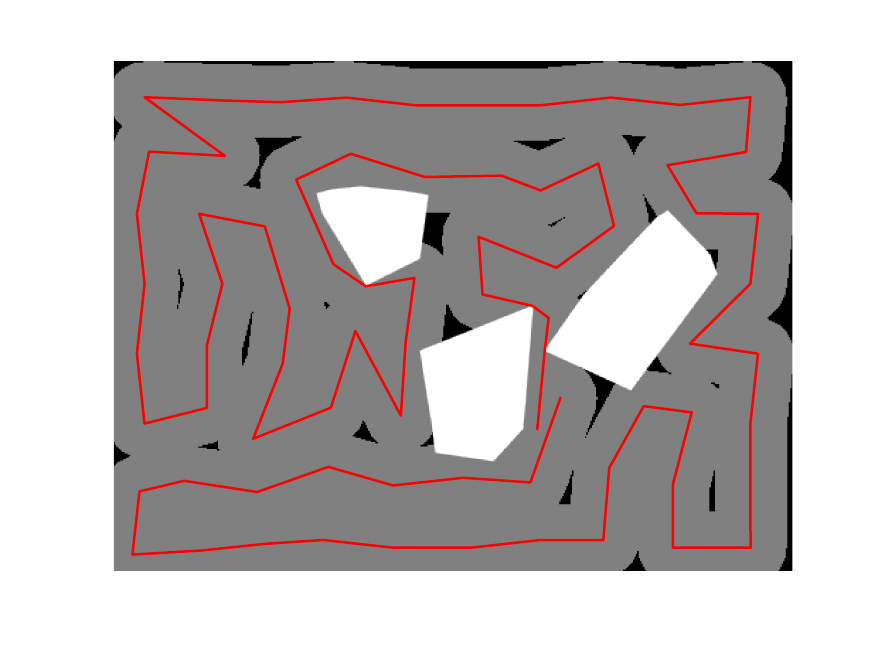}
  \caption{Representative trajectory: OCP path on a single map.}
  \label{fig:OCPExample}
\end{figure}

Figures \ref{fig:MSTsquareExample}--\ref{fig:OCPExample} display representative paths for the MST algorithm with a squared grid configuration, MST algorithm with a hexagonal grid configuration, the TSP algorithm, and the OCP approach. Although these particular trajectories come from one selected map, the subsequent performance metrics are aggregated from 10 runs with distinct randomly generated maps.

\subsection{Aggregate Performance Over 10 Samples}

Table \ref{tab:AggregateResults} summarizes the average uncovered area ratio (UAR, in \%), the average adjusted length of path (ALOP), and the average computation time (in seconds) for each algorithm over 10 distinct maps. UAR is computed as the percentage of total area left uncovered, while ALOP is calculated as
\[
  \frac{\text{(path length)} \times \text{(detection radius)}}{\text{covered area}}.
\]
The detection radius is assumed constant across algorithms.

\begin{table}[H]
\centering
\caption{Aggregated results over 10 samples. The order is TSP, MST-hex, MST-square, OCP.}
\label{tab:AggregateResults}
\begin{tabular}{lccc}
\hline
\textbf{Algorithm} & \textbf{Avg. UAR (\%)} & \textbf{Avg. ALOP} & \textbf{Avg. Time (s)}\\
\hline
TSP         & 4.3239  & 14.3604  & $1.3\times10^{-3}$\\
MST-hex     & 12.7821 & 13.7627  & $3.0\times10^{-4}$\\
MST-square  & 14.6655 & 13.2181  & $3.0\times10^{-4}$\\
OCP         & 3.3430  & 13.6339  & $1.2240\times10^{3}$\\
\hline
\end{tabular}
\end{table}

\begin{figure}[H]
  \centering
  \includegraphics[width=\textwidth,trim = 3cm 9cm 3cm 9cm,clip]{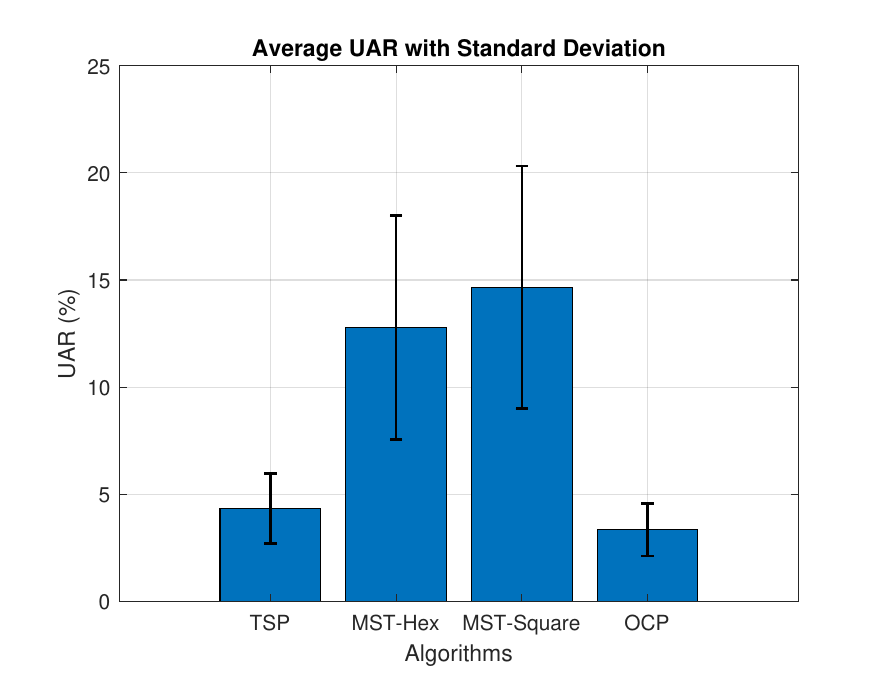}
  \caption{Average UAR (\%) with standard deviation for each algorithm.}
  \label{fig:UARComparison}
\end{figure}

\begin{figure}[H]
  \centering
  \includegraphics[width=\textwidth,trim = 3cm 9cm 3cm 9cm,clip]{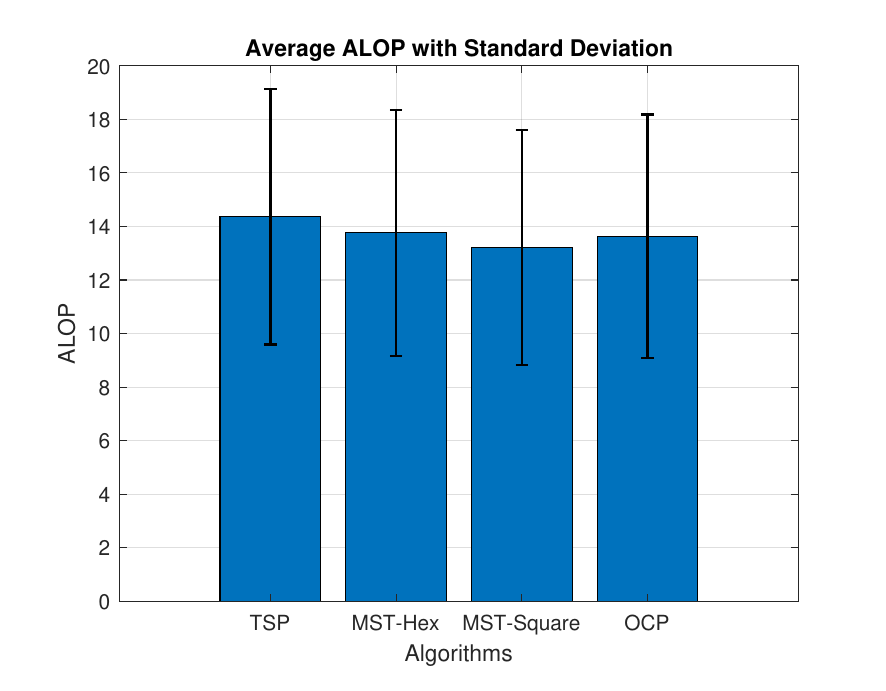}
  \caption{Average ALOP with standard deviation for each algorithm.}
  \label{fig:ALOPComparison}
\end{figure}

\begin{figure}[H]
  \centering
  \includegraphics[width=\textwidth,trim = 3cm 9cm 3cm 9cm,clip]{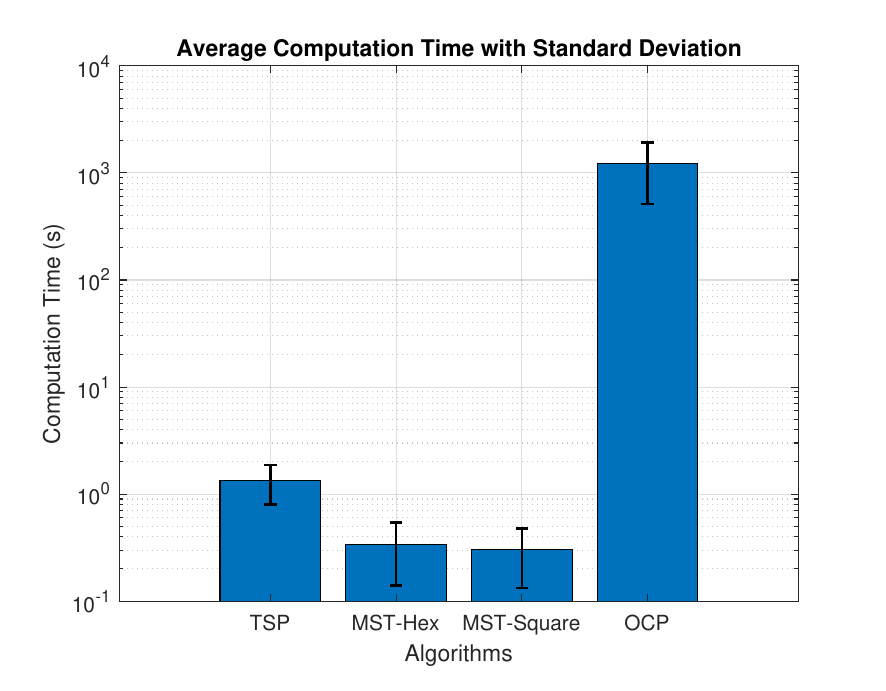}
  \caption{Average computation time (log scale) with standard deviation for each algorithm.}
  \label{fig:TimeComparison}
\end{figure}

Figures \ref{fig:UARComparison}--\ref{fig:TimeComparison} visualize these comparisons, with vertical lines indicating standard deviations.

\subsection{Discussion}

From Table \ref{tab:AggregateResults} and Figures \ref{fig:UARComparison}--\ref{fig:TimeComparison}, the OCP method achieves the smallest uncovered area (UAR $\approx 3.34\%$) but incurs the highest average computational time ($1.224\times10^3$~s). By contrast, MST-based algorithms and TSP require far less time, on the order of $10^{-4}$ to $10^{-3}$ seconds, though MST-square exhibits the largest UAR (14.66\%) among all four methods. TSP achieves a relatively modest UAR (4.32\%) but has the highest ALOP (14.36).

These results highlight a trade-off between thorough coverage (low UAR) and computational burden (time). Specifically, OCP delivers superior coverage at the expense of significant processing time. TSP offers a balanced compromise between coverage and runtime but yields a slightly higher path-length measure when normalized to the covered area and detection radius. Meanwhile, MST variations demonstrate excellent computational efficiency and can adapt to different sensor configurations (square vs.\ hexagonal), at the cost of covering less area.

In practice, the best choice depends on whether the mission prioritizes minimal uncovered area, shorter path distance, or lower computational overhead.

\subsection{Outer Polygon Extraction from Map Images}

Our approach automatically extracts an outer boundary polygon from map images to define the region of interest for coverage path planning. The segmentation process, based on color thresholding and subsequent contour extraction, produces a clean boundary delineating the area to be covered. This outer polygon is then used to compute a coverage path via an MST-based method, which scales efficiently and is therefore preferred over alternative approaches with higher computational costs.

\begin{figure}[H]
  \centering
  \includegraphics[width=\linewidth, trim = 6cm 12cm 6cm 11cm, clip]{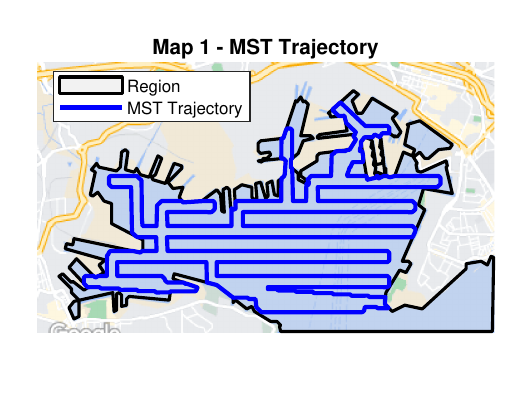}
  \caption{The Toulon map, with its extracted outer boundary and the computed MST-based coverage trajectory overlaid on the original image.}
  \label{fig:ToulonMST}
\end{figure}

\begin{figure}[H]
  \centering
  \includegraphics[width=\linewidth, trim = 4cm 9cm 4cm 8cm, clip]{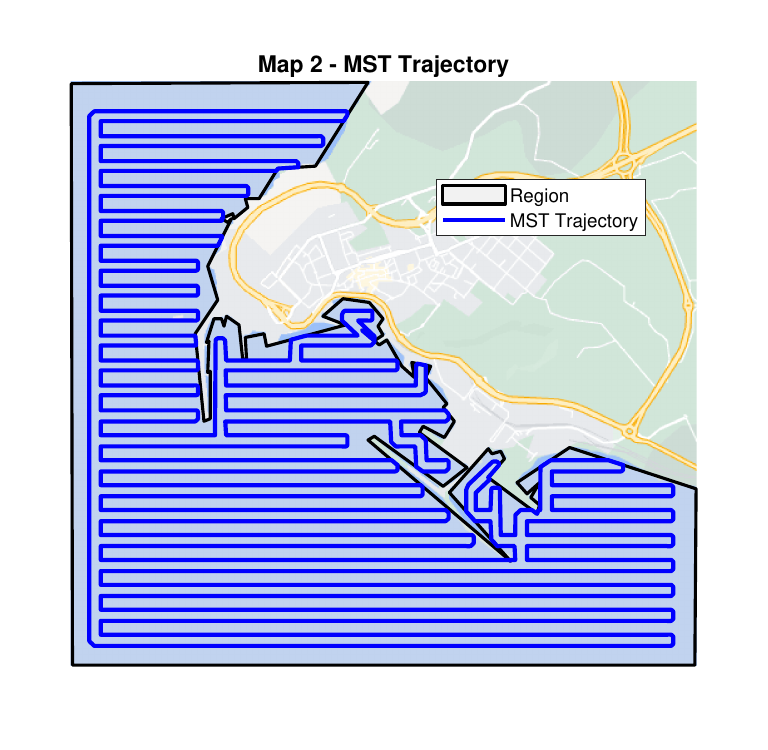}
  \caption{The Sines map showing the extracted boundary and the MST-based trajectory, demonstrating effective area coverage.}
  \label{fig:SinesMST}
\end{figure}

\begin{figure}[H]
  \centering
  \includegraphics[width=\linewidth, trim = 4cm 10cm 4cm 9cm, clip]{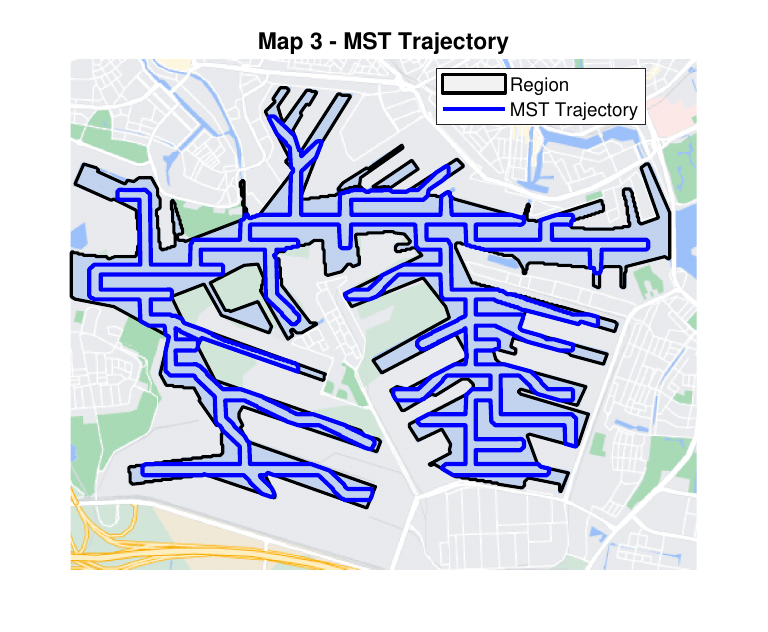}
  \caption{The Rotterdam map with the outer polygon and MST-based coverage path, ensuring all areas within the boundary are visited.}
  \label{fig:RotterdamMST}
\end{figure}

Figure~\ref{fig:ToulonMST} shows the result for the Toulon map, Figure~\ref{fig:SinesMST} for the Sines map, and Figure~\ref{fig:RotterdamMST} for the Rotterdam map. In all cases, the original map image is presented in the background; the extracted outer polygon is overlaid as a semi-transparent contour, and the corresponding MST-based trajectory is superimposed in blue.

These results confirm that our framework reliably extracts the region’s outer boundary from various map images, enabling efficient coverage path planning using the MST-based method.

	\showoutput
\section{Conclusions}
\label{chap:conclusion}

Trajectory generation is a crucial problem in the context of autonomous vehicle operations, particularly when extensive area coverage and efficient traversal are required. Coverage algorithms are designed to generate trajectories that maximize the probability of detecting specific targets while ensuring comprehensive coverage of the environment. This is often formulated as a generalized optimal control problem (OCP) with an appropriate detection model. 

To enable a fair comparison with alternative approaches found in the literature, the Travelling Salesman Problem (TSP) and Minimum Spanning Tree (MST) algorithms were implemented. These methods were evaluated against the OCP-based approach based on key performance criteria such as processing time, uncovered areas, and the total distance and duration of the generated trajectory. The results indicate that OCP is particularly useful when traversal time is constrained; however, it requires significantly higher processing time than alternative methods. Conversely, when maximizing coverage is the primary objective, TSP-based approaches perform better. Additionally, if a rapid solution is required, MST-based approaches yield lower computational costs.

\subsection{Future Work}

Applying a Voronoi-based coverage path planning approach could also yield promising results. Huang et al. \cite{huang2021novel} propose a method for path planning in irregular environments, such as those encountered in autonomous mowing applications. Their approach leverages Voronoi diagrams to segment the environment into distinct regions, allowing for systematic coverage while avoiding obstacles. Adopting a similar strategy in the context of vehicle navigation could enhance coverage efficiency and adaptability to complex terrains.


\bibliographystyle{plain}
\bibliography{references}
\end{document}